\documentclass{article}
\usepackage[square,numbers]{natbib}
\usepackage[utf8]{inputenc} 
\usepackage[T1]{fontenc} 
\usepackage{hyperref} 
\usepackage{url} 
\usepackage{booktabs} 
\usepackage{amsfonts} 
\usepackage{nicefrac} 
\usepackage{microtype} 
\usepackage{lipsum}
\usepackage{fancyhdr} 
\usepackage{graphicx} 
\graphicspath{{media/}} 
\usepackage{xcolor}
\usepackage[para]{threeparttable}

\usepackage{amsmath}
\usepackage{nccmath}
\usepackage{tabularx}
\usepackage{multirow}
\usepackage{amssymb}
\usepackage{subfig}

\usepackage{multicol}
\usepackage{array}
\usepackage{physics}
\usepackage{authblk}

\graphicspath{ {images/} }
\providecommand{\keywords}[1]
{
  \small	
  \textbf{\textit{Keywords---}} #1
}

\title{Forecasting Patient Flows with Pandemic Induced Concept Drift using Explainable Machine Learning

}
\author[1]{Teo Susnjak}
\author[1]{Paula Maddigan}
\affil[1]{School of Mathematical and Computational Sciences, Massey University, Auckland, New Zealand }
\begin{document}
\maketitle

\begin{abstract} 

Accurately forecasting patient arrivals at Urgent Care Clinics (UCCs) and Emergency Departments (EDs) is important for effective resourcing and patient care. However, correctly estimating patient flows is not straightforward since it depends on many drivers. The predictability of patient arrivals has recently been further complicated by the  COVID-19 pandemic conditions and the resulting lockdowns. 

This study investigates how a suite of novel quasi-real-time variables like Google search terms, pedestrian traffic, the prevailing incidence levels of influenza, as well as the COVID-19 Alert Level indicators can both generally improve the forecasting models of patient flows and effectively adapt the models to the unfolding disruptions of pandemic conditions. This research also uniquely contributes to the body of work in this domain by employing tools from the eXplainable AI field to investigate more deeply the internal mechanics of the models than has previously been done. 

The Voting ensemble-based method combining machine learning and statistical techniques was the most reliable in our experiments. Our study showed that the prevailing COVID-19 Alert Level feature together with Google search terms and pedestrian traffic were effective at producing generalisable forecasts. The implications of this study are that proxy variables can effectively augment standard autoregressive features to ensure accurate forecasting of patient flows. The experiments showed that the proposed features are potentially effective model inputs for preserving forecast accuracies in the event of future pandemic outbreaks.
 
\end{abstract}



\maketitle

\section{Introduction}\label{sec1}

Urgent Care Clinics (UCCs) and Emergency Departments (EDs) are frontline healthcare providers, provisioning continuous care for a range of acute medical presentations. For a significant proportion of patients, UCCs and EDs are the initial pathways en route towards subsequent healthcare services and are thus prone to significant periodic congestion. The effects of delayed treatments at EDs due to overcrowding have been well investigated and research shows that these delays have been associated with negative clinical outcomes  \cite{sudarshan2021performance}. Although UCCs typically cater to less acute medical conditions, studies have also noted that insufficient staff availability at these facilities could lead to serious conditions being overlooked \cite{Batal2001} in the presence of overcrowding.

Therefore it is important to implement strategies which assist with planning and the effective allocation of human resources to manage patient flows. Forecasting patient arrivals is one such strategy which can enable a greater optimisation of medical-service resourcing at facilities situated beyond primary care providers. Recently, there has been an increased academic interest in developing forecasting models for solving this problem. Forecasting of this kind can operate at different levels of granularity, such as estimating total daily patient volumes or even higher frequencies, like hourly patient arrivals. The purpose of this study is to develop daily patient-flow forecasting models using machine learning for two large UCCs providing services to patients experiencing sudden illness or accident-related injury.

Forecasting patient arrival volumes to a level of accuracy which makes the models useful is non-trivial due to numerous influencing factors as well as underlying stochastic processes. 
Producing reliable patient flow estimates has also been further complicated recently with the outbreak of the COVID-19 pandemic which has contributed an additional level of complexity due to the disruption to usual patient flow patterns \cite{hollander2021availablists,zhang2022forecasting}. These dislocations from otherwise stationary patterns in the dependent variable are termed \textit{concept drift}. This phenomenon is encountered when the variable being forecasted no longer corresponds to the original data used for creating the model, ultimately resulting in the deterioration of accuracies.  While these drifts can be sudden or gradual, ultimately a successful adaptive strategy depends on the ability to formulate an approach that captures the generative factors of the drifts as much as possible\cite{susnjak2012adaptive}. To date, no studies have yet emerged providing an analysis of how the concept drift arising from the COVID-19 pandemic and the resulting lockdown mandates have affected the patient-flow forecasting models, nor how the mitigation strategies can help adapt the forecasting models to increased volatility.

The current body of literature has primarily focused on forecasting solutions for EDs, with UCCs being largely overlooked; though the same drivers of patient demand and challenges apply to both UCCs and EDs. Most of the studies have concentrated on estimating total daily patient arrivals for the following day \cite{boyle2012predicting,xu2013modeling,navares2018comparing,whitt2019forecasting,rocha2021forecasting,harrou2022effective,zhang2022forecasting}. A few studies have investigated the feasibility of generating accurate forecasts as far as 30 days ahead \cite{marcilio2013forecasting,calegari2016forecasting} and some 90 days ahead \cite{maddigan2022forecasting}, which entails an even greater level of uncertainty and challenge due to compounding errors. For the most part, earlier studies relied on standard statistical methods for this type of modelling \cite{Batal2001,Jones2008,Champion_2007,Boyle2011}. Over time, machine learning approaches emerged \cite{xu2013modeling,zhang2022forecasting}, including deep learning  techniques \cite{sudarshan2021performance,rocha2021forecasting,harrou2022effective}. As the sophistication of the underlying algorithms has increased motivated by the attainment of high accuracies, the side effect has been a decrease in the interpretability of models. Little emphasis has been placed on clarifying the internal mechanics of the models. Examples of studies eliciting explanations of the models' reasoning underpinning their forecasts do not yet exist. 
Traditional time-series and simple regression methods have produced reasonable accuracies in previous studies. These approaches are still used and they continue to produce competitive results for forecasting patient flows. Research has shown that these techniques have excelled particularly in cases when the underlying data possessed consistent variations; however, it has been noticed that their accuracies become compromised when sporadic fluctuations are encountered \cite{harrou2022effective}. Approaches that combine the advantages of traditional approaches with those of machine learning have not yet emerged in this domain.

The features used for the forecasting models in the prior works have chiefly relied on autoregressive (previous or lagged values) and various calendar variables \cite{Jones2008,Wargon395,boyle2012predicting,calegari2016forecasting,rocha2021forecasting,harrou2022effective} with some studies using variables indicating school holidays as well. There has been a recent trend to also incorporate weather-related variables \cite{calegari2016forecasting,sudarshan2021performance,zhang2022forecasting, petsis2022forecasting}, while air quality and pollution information have been a subject of some studies\cite{sahu2014hierarchical,navares2018comparing}.

The aim of this study is to present a machine learning solution to forecasting total daily patient flows 7-days ahead in order to facilitate resource planning strategies at UCC facilities. This work particularly aims to investigate the feasibility of a range of real-time proxy variables to contribute to improving the overall model accuracies, as well as to employ a range of eXplainable Artificial Intelligence (XAI) techniques to both bring interpretability into the model behaviour and to extract additional insights. Additionally, this research seeks to develop mechanisms for enabling the models to adapt to the disruptive conditions of the recent COVID-19 pandemic, with learnings that can be applied to similar scenarios in future.

\subsubsection*{Contribution}

A key contribution of this research is the use of novel, near real-time\footnote{The term real-time has different meanings depending on a domain. Real-time in this context is defined by data that has been updated within a 24-hour window.} proxy features for forecasting patient flows at UCCs, with an emphasis on creating strategies for adapting the forecasting models to underlying concept drifts due to the pandemic conditions, with broad applicability to EDs as well. The features used in the study consist of Google Trends search terms, pedestrian traffic, weather information, reports of prevailing incidence levels of the seasonal flu and pandemic Alert Level variables. This research also uniquely addresses a gap in this domain associated with the use of uninterpretable black-box models. This work uses the latest XAI tools for depicting the behaviour of the models both at high-level with respect to feature importances, and at a finer granular level to demonstrate how the models explain the reasoning behind their forecasts. To the best of our knowledge, this work is distinctive within this problem domain in demonstrating comprehensively how these tools can be leveraged. This study is broad in the use of a wide range of machine learning algorithms and achieves competitive accuracies with prior works.    

\section{Related work}

The vast majority of prior research in forecasting patient flows has focused on EDs. Some prior studies \cite{Wargon395} have used patient flow forecasting literature from EDs and UCCs interchangeably and thus we follow the same pattern with the understanding that approaches that have been effective in the ED domain are transferable to the UCC context, and vice versa. 

We review published literature which is most comparable to our work in terms of studies that have predominately investigated the forecasting of total daily volumes of patient arrivals. We also focused on studies which reported their forecasting accuracies in terms of the Mean Absolute Percentage Error (defined in Equation \ref{MAPE}) which to some degree enables comparisons across different studies. Forecast horizons in terms of the number of days ahead being estimated are relevant for this study, and these are also highlighted and summarised where possible in Table \ref{table:litsummary}. 


Traditional regression and autoregressive modelling dominated the earliest research into forecasting patient demand at EDs. \citet{Batal2001} successfully used multiple linear regression (MLR) to predict patient flows by incorporating a selection of calendar variables like the day of the week, month, season, and holiday flags. \citet{Jones2008} subsequently used the same set of variables together with MLR as a benchmark model against comparisons with autoregressive ARIMA and SARIMA models for predicting daily patient presentations. Their work fundamentally differed in that they extended their forecast horizon to 30 days ahead. 
Meanwhile, \citet{Boyle2011} focused on next-day ED patient flow forecasting, while using MLR, ARIMA and Exponential Smoothing (ES) models. In contrast, both \citet{Champion_2007} and \citet{Aboagye_Sarfo_2015} altered the frequency from daily, to predicting the total monthly patient demand. Both studies relied on  ES with ARIMA modelling methods, while \citet{Aboagye_Sarfo_2015} also integrated these algorithms with experiments using VARMA.

The focus of subsequent works explored the viability of forecasting at differing frequencies together with longer forecasting horizons. One-step-ahead hourly, daily and monthly ED patient flow forecasting models with the use of calendar information were developed by \citet{boyle2012predicting}. Their work also experimented with standard approaches like ES, ARIMA and MLR, concluding that the errors increased and thus the forecasting became less reliable as the forecast granularity became finer. \citet{marcilio2013forecasting} returned to forecasting daily patient flows, exploring 7 and 30-day ahead predictions with the help of calendar as well as climatic variables. They expanded the range of techniques used to Generalized Linear Models (GLM), Generalized Estimating Equations (GEE) and Seasonal ARIMA (SARIMA).  
Whilst recognising the importance of forecasting further out into the future, \citet{calegari2016forecasting} addressed predicting total daily ED patient flows over 1,7, 14, 21 and 30-days ahead, while also relying on the calendar and climatic data. They also used standard SARIMA as in previous studies, but also included Seasonal ES (SES), Seasonal Multivariate Holt Winter’s ES (HWES) and Multivariate SARIMA. 
 
\citet{xu2016hybrid} proposed a method that combined ARIMA and Linear Regression (ARIMA-LR) to predict daily ED patient flows for both the next day as well as 7-days ahead forecasting. 

Given the effectiveness of traditional approaches in this domain, more recent studies have continued to leverage these techniques like \citet{carvalho2018assessment}, who experimented with forecasting at different levels of granularity. They forecasted total patient flows on a next-week and next-month basis using ARIMA and ES, while \citet{whitt2019forecasting} forecasted next-day total patient flows using SARIMAX using also calendar and climatic variables.


\begin{table}[htb]
\centering
\setlength\tabcolsep{1pt}
\caption{Summary of literature predicting total daily patient flows in EDs, highlighting forecasting horizons, best algorithms and the accuracies achieved.}  
\begin{tabular} 
{ p{0.25\textwidth} 
>{\centering}p{0.1\textwidth}
>{\centering}p{0.12\textwidth}
>{\centering}p{0.12\textwidth} 
>{\centering}p{0.16\textwidth}
>{\centering \arraybackslash}p{0.21\textwidth}}

\toprule
Study& Year&	Forecast periods (days)& Features &	Best algorithm& 	MAPE\\
 \midrule
\citet{boyle2012predicting} &\citeyear{boyle2012predicting} &1&4	 &	ARIMA,ES,OLS& 	7.0\%\\
\citet{xu2013modeling} &\citeyear{xu2013modeling}&	 1&	7 &	ANN& 6.8\% - 7.3\%\\
\citet{marcilio2013forecasting} &\citeyear{marcilio2013forecasting} &	7	 &15	&GLM& 7.6\%\\
&	&	30&	 &	GLM& 9.7\%\\
\citet{xu2016hybrid} &\citeyear{xu2016hybrid}&1&	31 &	ARIMA-LR& 	6.5\%\\
&	&	7 &		&ARIMA-LR& 9.6\%\\
\citet{calegari2016forecasting}& \citeyear{calegari2016forecasting}&	1 &	5	&SES& 	2.9\%\\
&	&	7& &		SES& 10.7\%\\
&	&			14& &		SES& 10.7\%\\
&	&			21& &		SES& 	11.4\%\\
&	&			30& &		SES& 11.7\%\\
\citet{navares2018comparing}& \citeyear{navares2018comparing}&	1& 13 &		ARIMA& 8.1\% - 12.3\%\\
\citet{whitt2019forecasting} &\citeyear{whitt2019forecasting}&		1& 57&		SARIMAX& 8.4\%\\
\citet{rocha2021forecasting} &\citeyear{rocha2021forecasting}&	1& 15&		LSTM& 	4.2\%\\
\citet{sudarshan2021performance} &\citeyear{sudarshan2021performance}& 	3& 10&		CNN& 9.2\%\\
&	&	7& &		LSTM& 	8.9\%\\
\citet{harrou2022effective}	&\citeyear{harrou2022effective}&	1 &	7	&DBN& 	4.1\%\\
\citet{zhang2022forecasting}& \citeyear{zhang2022forecasting}&	1& 29&		SVR& 	8.8\%\\

\citet{petsis2022forecasting} &\citeyear{petsis2022forecasting}&	1& 38&		XGBoost& 	6.5\%\\
 &&	2& &		XGBoost& 	6.9\%\\
\bottomrule
\end{tabular}
\label{table:litsummary}
\end{table}

Though traditional time-series and regression approaches possess more assumptions and stricter requirements that are harder to satisfy, they do continue to generate effective solutions for forecasting patient flows. In contrast, non-parametric machine learning approaches avoid many of the rigid assumptions required by traditional statistical methods and therefore have some added flexibility accompanied by a proclivity to overfit. 
Over the past decade machine learning solutions in this domain have started to emerge and have consistently produced competitive accuracies, while also exceeding those of traditional approaches in some cases \cite{xu2013modeling,rocha2021forecasting}. 
Machine learning methods generally have the ability to capture non-linear relationships in the data and thus hold the potential to produce models which arguably better represent the complex and dynamic nature of patterns in this domain \cite{zhang2022forecasting}.

\citet{xu2013modeling} was one of the earliest works using machine learning in this domain where they applied Artificial Neural Networks (ANN) in conjunction with variables describing seasonal influenza epidemics, as well as calendar and climatic data in order to forecast one-day ahead total daily patient flows. 
We see the emergence of ensemble-based techniques like Random Forest (RF) and Gradient Boosting Machines (GBM), together with ANN for forecasting daily respiratory and circulatory-related ED admissions in work by \citet{navares2018comparing}. In this study, the authors incorporated environmental and bio-meteorological variables into the models that captured air quality indicators, and ultimately concluded that machine learning methods did outperform ARIMA when the models were combined using Stacking.

We are now beginning to see an emergence of a broader range of proxy variables being used, together with deep learning approaches and some early signs of XAI techniques entering this domain.  \citet{Vollmer_2021} proposed patient flow models forecasting 1, 3 and 7-days ahead while also using RF and GBM, together with a k-Nearest Neighbours (kNN) regressor. They expanded the range of proxy variables used in their models and included weather, school and public holidays, and seasonal pattern data, alongside large scheduled events which tended to see an influx of patients. Additionally, they also utilised Google search data for the keyword “flu”.  
Long Short-Term Memory (LSTM) was used by \citet{rocha2021forecasting} where they performed a wide range of comparative experiments with those of traditional ES and SARIMA approaches and other machine techniques like Autoregressive Neural Networks (AR-NN) and XGBoost. The authors concluded that LSTM produced the best accuracies for next-day forecasts of total daily patient flows. \citet{sudarshan2021performance} likewise used LSTM for predicting 3 and 7-day ahead total daily patient flows. They included  Convolutional Neural Networks (CNN) into their suite of algorithms and compared the results against RF. The authors also confirmed the effectiveness of deep learning methods in this domain by demonstrating that LSTM was most accurate for the 7-day forecasting, while CNN was more suited for 3-day ahead forecasting in their setting. 

\citet{harrou2022effective} expanded the range of deep learning techniques in their work by including Deep Belief Networks (DBN), Restricted Boltzmann machines (RBM), Gated Recurrent Unit (GRU), a combination of GRU and Convolutional Neural Networks (CNN-GRU), CNN-LSTM, as well as the Generative Adversarial Network based on Recurrent Neural Networks (GAN-RNN). They designed their models to predict total daily patient flows one-day ahead, finding that the best accuracy was achieved using DBN. However, research also demonstrated that deep learning methods do not always deliver the best accuracies. 

The study by \citet{zhang2022forecasting}, contrasted LSTM with standard machine learning methods like kNN, Support Vector Regression (SVR), XGBoost, RF, AdaBoost, GBM and Bagging, in addition to traditional methods like OLS and Ridge Regression. This study found that it was SVR which outperformed all other algorithms based on models predicting next-day patient flows while using calendar and meteorological variables.

Ensemble-based techniques leveraging GBM such as XGBoost and LightGBM continue to increase in prominence. \citet{guyeux2022predict} used both algorithms together with RF and Lasso for hourly forecasting of patient flows. Similar to our study, the authors expanded the use of variables for forecasting beyond using standard calendar and meteorological variables. Instead, the study augmented these variables with road traffic information and epidemiological data reporting the current number of cases of influenza, chicken pox and acute diarrhoea. All in all, the study used 712 explanatory variables for forecasting and concluded that XGBoost was best performing in their setting. Meanwhile, another recent study bearing parallels with our research was conducted by \citet{petsis2022forecasting}, who focused on predicting daily patient flows 1 and 2 time periods ahead. This work is reported to be the first study that has started to incorporate XAI techniques into illuminating the behaviour of the underlying models. This work primarily considered the high-level interpretability of the models which included the analysis of feature interactions, but the reasoning behind the forecasts for specific days was not considered. The study again used Gradient Boosting in the form of XGboost, relying on standard calendar and weather variables which amounted to a total of 38 variables.

\subsection{Summary}

\begin{figure}[hbt]
\begin{tabular}{cc}
\subfloat(a){%
\includegraphics[scale=0.37]{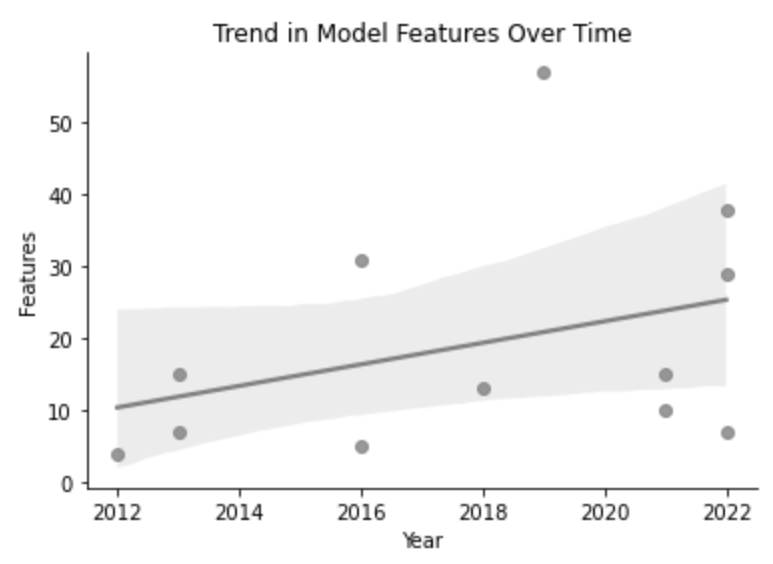}} & 
\subfloat(b){%
\includegraphics[scale=0.3]{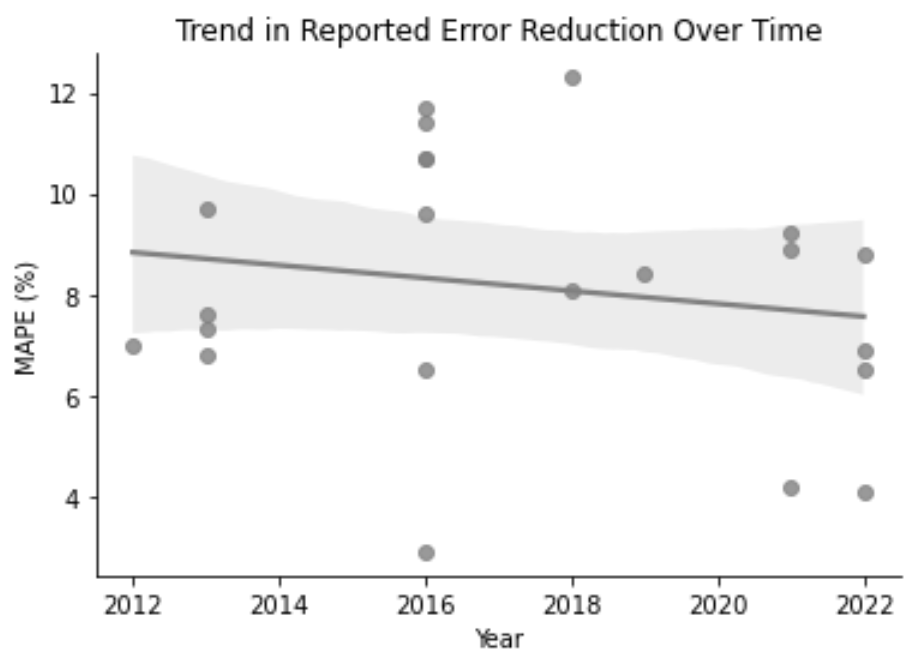}
} \\
\end{tabular}
\caption{Trend in the reported accuracy of forecasting patient flows over time according to MAPE.}
\label{Fig:mape_existing}
\end{figure}

It is natural to see that traditional techniques for forecasting patient flows were the most prominent approaches in earlier research. The emerging trend within the existing literature points in the direction of increasing use of machine learning algorithms, particularly those with more inherent complexity. This can be explained by a need to use algorithms that are capable of extracting the signal from the noise as the size of the datasets and that of the various features have been increasing over time (Figure \ref{Fig:mape_existing}a). Though machine learning approaches are achieving a widespread uptake, traditional methods continue to be used as well and generate competitive accuracies; however, they are now more often used as benchmarking models. 

Thus far, the literature indicates that amongst the researchers autoregressive, calendar and holiday variables, together with weather and bio-meteorological features remain popular model inputs for forecasting. Some recent investigations into the usefulness of the proxy variables like internet search terms and live influenza tracking indicators are emerging. Due to this and an increasing number of features being used, the suitability of non-parametric algorithms becomes evident. Traditional approaches possess advantages which can still be leveraged. However, no studies exist that attempt to combine both machine learning and traditional techniques into a single model that is able to rely on the advantages that both approaches offer. Additionally, the literature shows that there is an emergence of an ever more diverse range of machine algorithms being used in this domain; however, there is no clear algorithm which stands out as the best-performing. 

Achieving high accuracy in forecasting patient flows has been a central focus of most prior studies. Figure \ref{Fig:mape_existing}b depicts the trend in the reported forecasting accuracy over time. The figure reveals a gentle downward slope indicating a gradual reduction in error (reported as the Mean Absolute Percentage Error - MAPE) values over time. It should be noted that only a single study used datasets covering the COVID-19 period which exhibited sudden and recurring concept drifts in the underlying data patterns, which result in deteriorations of accuracy. Therefore, the overall picture presented in the figure is arguably optimistic until newer studies emerge and are published which include results from this period.

Finally, the increased usage of more complex machine learning algorithms is accompanied by the generation of uninterpretable "black box" models. A clear gap in the literature for patient flow forecasting exists in the absence of analyses which consider both the high-level mechanics of model behaviour as well as the interrogation of the models' reasoning into specific forecasts. No study as yet using machine learning has explored the internals of the models to the fullest extent.

In view of the existing literature, the research questions (RQ) addressed in this study are:
\begin{itemize}
\item (RQ1) Which machine learning models attain the highest improvements over the benchmark strategies for short-term forecasting of daily patient arrivals at UCCs? Can machine learning and traditional statistical approaches be combined to produce higher accuracies?

\item (RQ2)  Can proxy variables improve forecast accuracies and adapt the models to the concept drift caused by the COVID-19 conditions? Which variables are most effective at generally improving the forecasting accuracies of daily patient demand seven days ahead? 

\item (RQ3) How can greater interpretability and explainability of forecasting models be achieved? 

\end{itemize}

\section{Methodology}

\subsection{Setting} The data were sourced from Shorecare\footnote{https://www.shorecare.co.nz/} which owns and operates the clinics in this study. The Smales Farm clinic provides 24-hour care services, while the Northcross clinic is reduced to after-hours care. The clinics treat standard low acuity cases and provide x-ray and fracture clinics as well as facilities for complex wound management. The Smales Farm clinic is the sole 24-hour UCC servicing a population of approximately one-quarter of a million and is situated within one kilometre of a major hospital whose ED treats $\sim$46,000 patients annually.

\subsection{Patient Flow Dataset} 
\begin{figure*}[htb]
\centering
\includegraphics[scale=0.44]{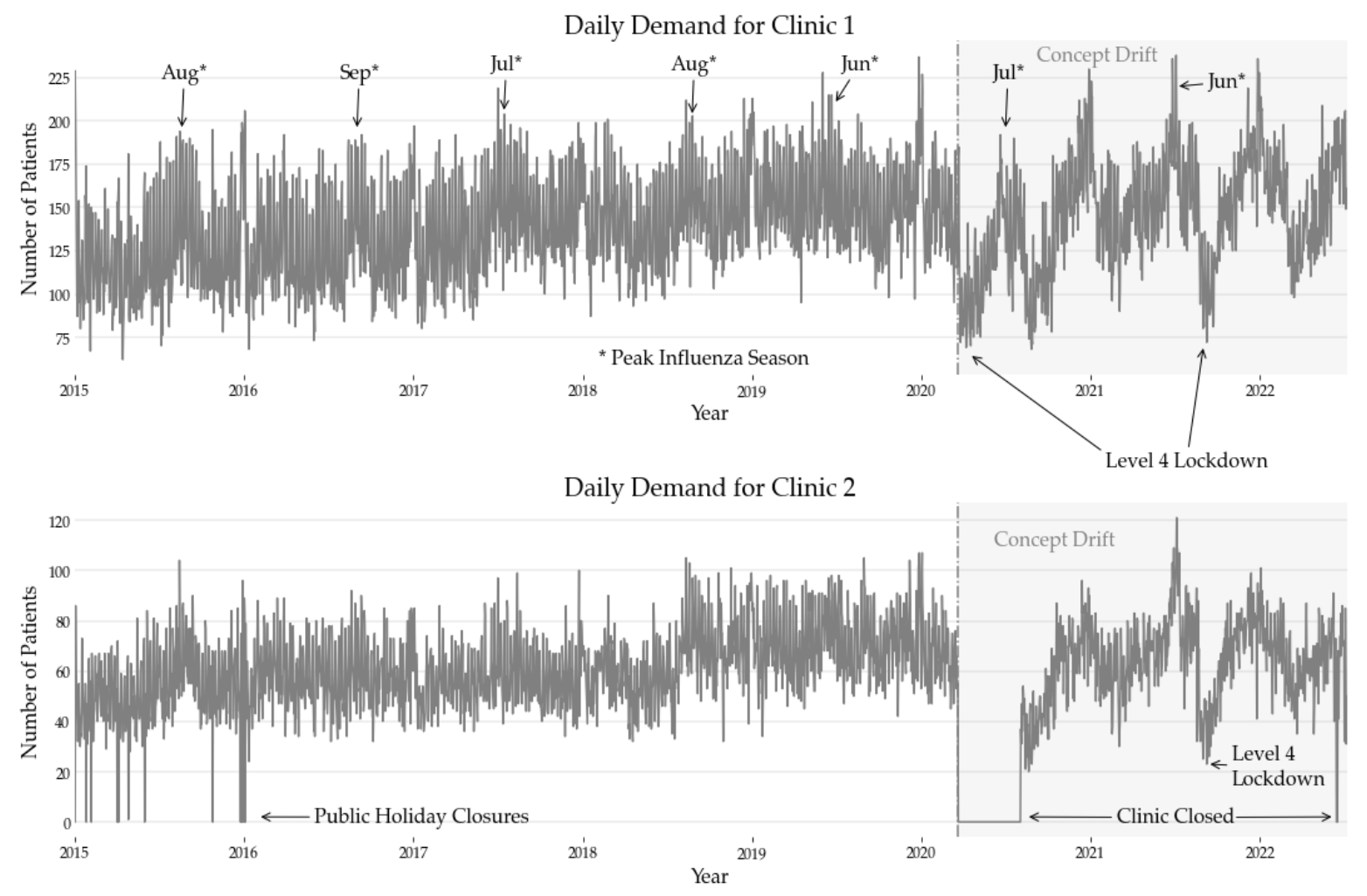}
\caption{Daily patient demand at both UCC clinics ranging from 2015 - 2021}
\label{figure:ts}
\end{figure*}

Models were designed for predicting daily arrivals seven days ahead. The dataset recorded patient presentations spanning 11 years from 2011 through to 2022. Figure \ref{figure:ts} shows the characteristics of the patient flows on a selection of the dataset for both clinics. Seasonal patterns indicating increases in patient arrivals are visible during the winter months of the Southern Hemisphere (June-August). This can be explained by increased presentations of  Influenza and other respiratory-related illnesses. The beginning of the period affected by strong concept drifts is highlighted and deviations from the prior patterns can be observed. Additionally, the exact points corresponding to the most stringent mandates concerning COVID-19 pandemic lockdowns are highlighted, as well as periodic partial closures of the smaller clinic during portions of this period.  

\begin{figure}[hbt]
\centering
\includegraphics[scale=0.45]{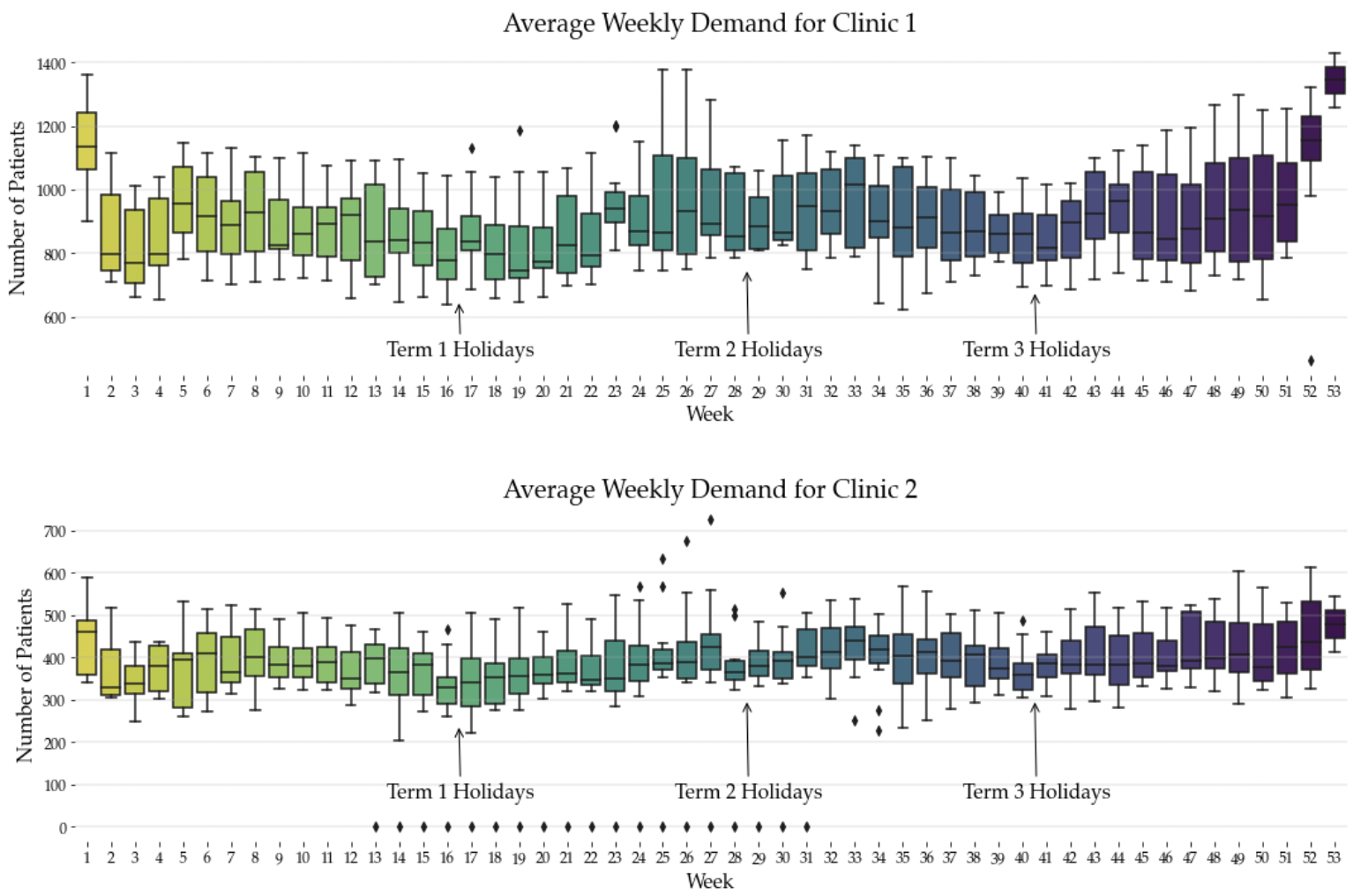}
\caption{Average patient demand across both UCCs for all years by week number.}
\label{figure:weekly}
\end{figure}

An aggregation of the data by week number across all years can be seen in Figure \ref{figure:weekly}. There is generally a reduction in demand during the school holidays, with peaks in patient arrivals coinciding with the Winter months (mid-year), and the end of the year when the local general practitioners (GPs) are closed for holidays.

\begin{figure}[hbt]
\centering
\includegraphics[scale=0.45]{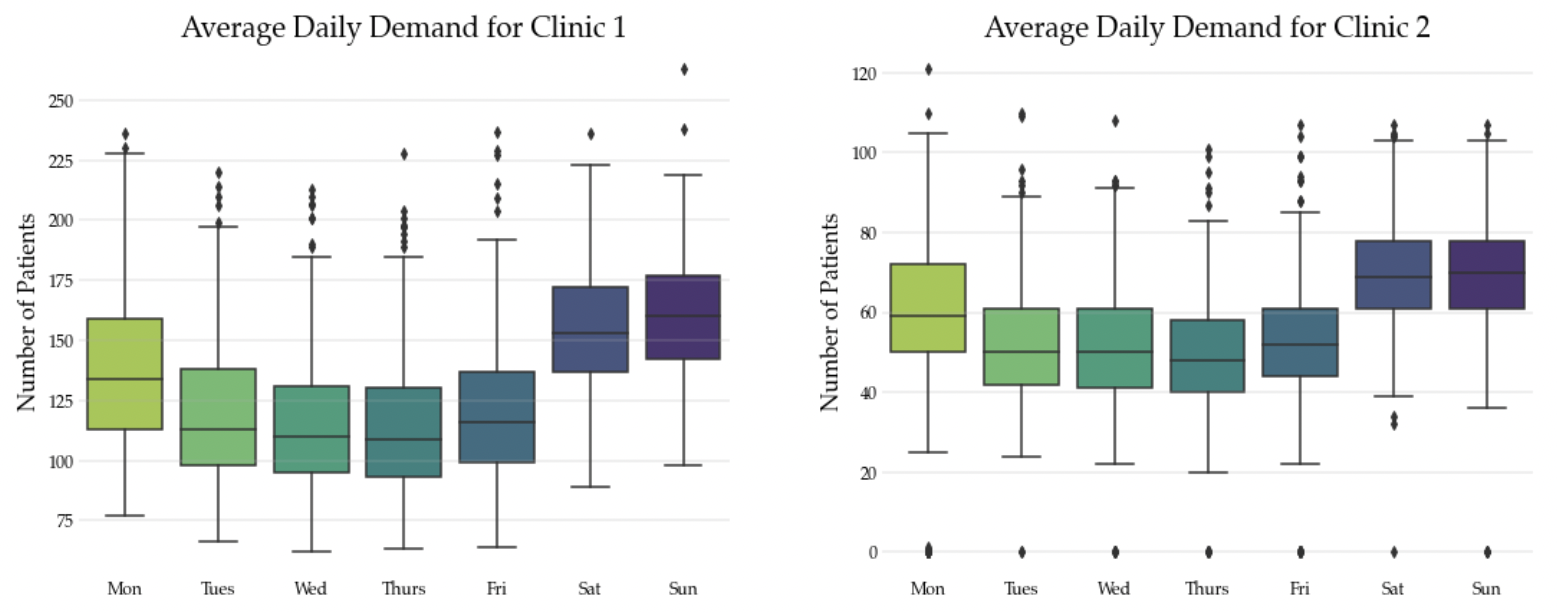}
\caption{Average daily patient demand by day of the week across both UCCs.}
\label{figure:dow}
\end{figure}

Figure \ref{figure:dow} shows average patient arrival patterns across both clinics by day of the week. A peak in patient presentations tends to occur during the weekend when GPs are closed, which is followed by a decrease in patient flows at the start of the working week, which then reaches minima by mid-week.

\subsection{Proxy Data and Features} The models use a mixture of autoregressive and proxy features. Since the data exhibits pronounced seasonal and weekly cyclical patterns, we used autoregressive features representing previous patient arrival values from 7, 14, 364, 728 and 1092 days before as predictors.  Additionally, the week number in the year was used to provide information describing the seasonal trends as well as recurring school holiday patterns. 
A public holiday flag was devised which ensured that the elevated demand driven by these days was represented. 

A strategy was devised to handle the pandemic-related concept drift by using a set of variables which had the ability to capture the changes in the underlying conditions. Primarily, we created a variable that represented the severity of the legally mandated COVID-19 restrictions\footnote{This expressed either the COVID-19 Alert Level system \cite{NZG2022}  or subsequent Traffic Light mappings \cite{NZG2022a} defined by the New Zealand Government which were later introduced.} prevailing within the clinics' locality. 

Other variables were used with the assumption that they would capture both the signal indicating that concept drift was ocurring, and that they would also contribute to the general improvement of the overall models by being effective proxies for patient demand.  For this, we used \citet{GoogleTrends} data which provides information on the frequency of Google search keywords over a given time period and region. We used this data as a proxy variable in order to capture potential increases in various symptoms which may trigger higher patient arrivals. The selected keywords were \textit{flu, headache, sick, chest pain} and \textit{cough} being both generic as well as relevant to the COVID-19 pandemic. Google Trends data is accessible at weekly aggregate results provided that the search criteria fall within 5 years.

Since lockdown mandates affect population movements, we included pedestrian foot traffic data as additional proxy variables in the models. We used data from \citet{City2022}, which provides information on the number of pedestrians walking past approximately 20 cameras in the Central Business District (CBD) of Auckland each day, with data stretching back to 2012.

Similar to previous studies, we also integrated information on the prevailing weather conditions. We sourced the data using APIs from \citet{VisualCrossings}. Our exploratory analysis indicated that the most effective feature from the suite of possible weather variables was the \textit{'feels like'} indicator that combines temperature, wind chill and heat index values.

We also incorporated the data from FluTracker \cite{FluTracking2022}. FluTracker monitors prevailing levels of Influenza in New Zealand using surveys. Approximately 30,000 people report weekly symptoms of fever or cough being experienced. The New Zealand FluTracker data is available only from 2018 onwards, with weekly reports provided online.  The data also has missing values which coincide with the summer months when the prevalence of the illness is usually low. The data is presented as the percentage of respondents exhibiting symptoms each week.

Table \ref{table:FeatureDesc} lists the names of all the variables used in various figures together with their descriptions. Appendix \ref{appendix:proxy_vars} shows end example of the proxy feature values versus patient flows in Figure \ref{figure:proxy_ped} to Figure \ref{figure:proxy_trends}.

\begin{table}[hbt]
\centering
\caption{Feature names as they appear in figures and their descriptions.}   
\begin{tabular} {p{0.15\textwidth} p{0.8\textwidth}@{}}
\hline 
Feature name & Description
\\
\hline 
lag7d  & Autoregressive 7-day lag, the value from one week prior\\
lag14d & Autoregressive 14-day lag, the value from two weeks prior \\
lag1 & Autoregressive 364-day lag,  the value from one year prior \\
lag2 & Autoregressive 728-day lag,  the value from two years prior \\
lag3 & Autoregressive 1092-day lag, the value from three years prior \\
public\_holiday & Public Holiday Indicator (0/1)\\
week & Week Number ranging from 1 - 53\\
ped\_count & Auckland CBD foot traffic  \\
flu\_percent &  weekly reports on Influenza prevalence in New Zealand \\
covid\_level & COVID-19 Alert Level ranging from 1-4, or Traffic Light (Green=1; Orange/Red=2)\\
trends & Google Trends normalised frequency of term searches of \textit{flu, headache, sick, chest pain} and \textit{cough} as a single signal \\
feels\_like & A combination of temperature, wind chill and heat index values \\

     \hline
\end{tabular}
 
\label{table:FeatureDesc}
\end{table}

\subsection{Benchmark models} We evaluated the efficacy of the proposed models against several benchmark models. The primary benchmark model replicated the current in-house strategy employed by the clinics' administrators to estimate patient arrivals. This approach used the patient arrival numbers from the same period in the previous year plus an additional 5\% to account for an increasing trend in total volumes. 

The second benchmark model is a Persistence Model. It is essentially a Random Walk technique \cite{pearson1905problem} which makes the forecasted value the same as that of the identical day in the previous year. The third benchmark model was ARIMA \cite{box1976time}, created with auto-tuning. The final benchmark model consisted of an enhanced version of the Persistence Model which forecasted a value for a given day at a point in time \textit{t} to be a mean of weighted autoregressive features consisting of values from \textit{t}-7, \textit{t}-14,  \textit{t}-364, \textit{t}-728 and \textit{t}-1092 days. 

\subsection{Algorithms} We used ten statistical and machine learning algorithms in order to generate candidate models. These consisted of: Random Forest (RF) \cite{breiman2001random}, Voting \cite{scikit-learn}, Stacking \cite{wolpert1992stacked}, Ridge Regression \cite{hoerl1970ridge}, Support Vector Machines Regression (SVR) \cite{drucker1996support}, Kernel Ridge Regression (KRR) \cite{cristianini2000introduction}, K-Nearest Neighbour Regression (kNN) \cite{cover1967nearest} as implemented in Scikit-learn \cite{scikit-learn}, CatBoost \cite{catboost}, Prophet \cite{taylor2018forecasting} and an Averaging Model. Table \ref{algorithms} lists all the algorithms as well as the benchmark models with their brief descriptions, together with their hyperparameter values where relevant.

\begin{table}[p]

\caption{Summary of the benchmark models and algorithms used in this study, together with hyperparameter settings where applicable.}\label{algorithms}
\fontsize{8pt}{9pt} 
\selectfont
\begin{tabular} 
{p{0.17\textwidth} 
p{0.77\textwidth}@{}}

\toprule

Method  & Description  \\ 
\midrule
Benchmark & Current estimation method used in-house by the clinics which forecasts patient demand for a given day to be 5\% higher than that of the same day in the previous year. \\
Persistence Model & A benchmark model implemented as a Random Walk \cite{pearson1905problem} method with the forecast being the same as the value for the same period of the previous year \\
Enhanced Persistence Model  & An optimised benchmark model that made forecasts based on the weighted mean value of autoregressive values in respect to time \textit{t} with time lags of \textit{t}-7, \textit{t}-14,  \textit{t}-364, \textit{t}-728 and \textit{t}-1092. The weightings were optimised through an empirical approach and set as [5, 4, 3, 2, 1] respectively for each autoregressive feature, with features representing recency being allocated greater importance.  \\

ARIMA  & Traditional autoregressive statistical technique, predicting future values based on past values. \\

kNN Regression  & A non-parametric algorithm that bases its predictions on the principle of proximity, producing a forecast that is an aggregation of \textit{k} nearest observations with respect to the characteristics of the data point in question.  \\

Ridge Regression  & A technique that creates a parsimonious model which shrinks the coefficients towards zero using L2 regularization. The resulting models generally reduce the variance resulting in an improved mean-squared error. \\

Support Vector Machines Regression & SVR is an extension of SVMs. It uses a pre-defined kernel function to transform the data from a non-linear space to a higher dimension in order to find an approximate fit that satisfies a pre-determined error margin. To that end, the objective function of SVR is to reduce the coefficients rather than the error term (epsilon). SVRs are particularly effective on smaller datasets and are more robust to outliers. \\

Kernel Ridge Regression  & Kernel ridge regression extends Ridge Regression with the integration of the kernel trick technique from SVR. It differs to SVR in that it uses the squared error loss as opposed to the epsilon-insensitive loss in SVR, combined with l2 regularization. \\

Prophet  & Auto-tunable, additive forecasting model with the ability to handle non-linear trends using yearly, weekly, and daily seasonality with capabilities to integrate effects from holidays, having robustness to dislocations in trend.  \\

Random Forest Regression & Ensemble-based algorithm consisting of decision trees whose outputs are combined. Each decision tree is induced based on random feature subsets, resulting in an uncorrelated forest of trees. The combined accuracy of the forest results in a higher fidelity than that of any individual tree. \\

CatBoost &  CatBoost is an ensemble-based algorithm that generates gradient-boosted decision trees. During training, successive trees are induced with a reduction in loss. The size of the ensemble is preset by defining the maximum number of trees as a parameter.\\

Voting Regressor  & Ensemble-based meta-estimator. Combines machine learning and traditional time-series approaches. Initially generates models for the underlying base regressors: Prophet, CatBoost, Random Forest and ARIMA. It then combines the outputs of these algorithms for the final forecast using a weighted combination scheme.  \\

Averaging Model &  This algorithm was a customised version of the Voting Regressor which combined the outputs of five algorithms (Prophet, CatBoost, Random Forest, Voting and Stacking) but discarded the highest and lowest predictions in the calculation of each prediction. \\

Stacking & An ensemble-based meta-estimator which models the forecast outputs of the underlying base estimators (Prophet, CatBoost and Random Forest) using an overarching regressor whose output constitutes the final forecast.  \\

\bottomrule
\end{tabular}
\end{table}

\normalsize

\subsection{Testing approach}
\begin{figure*}[htb]
\centering
\includegraphics[scale=0.35]{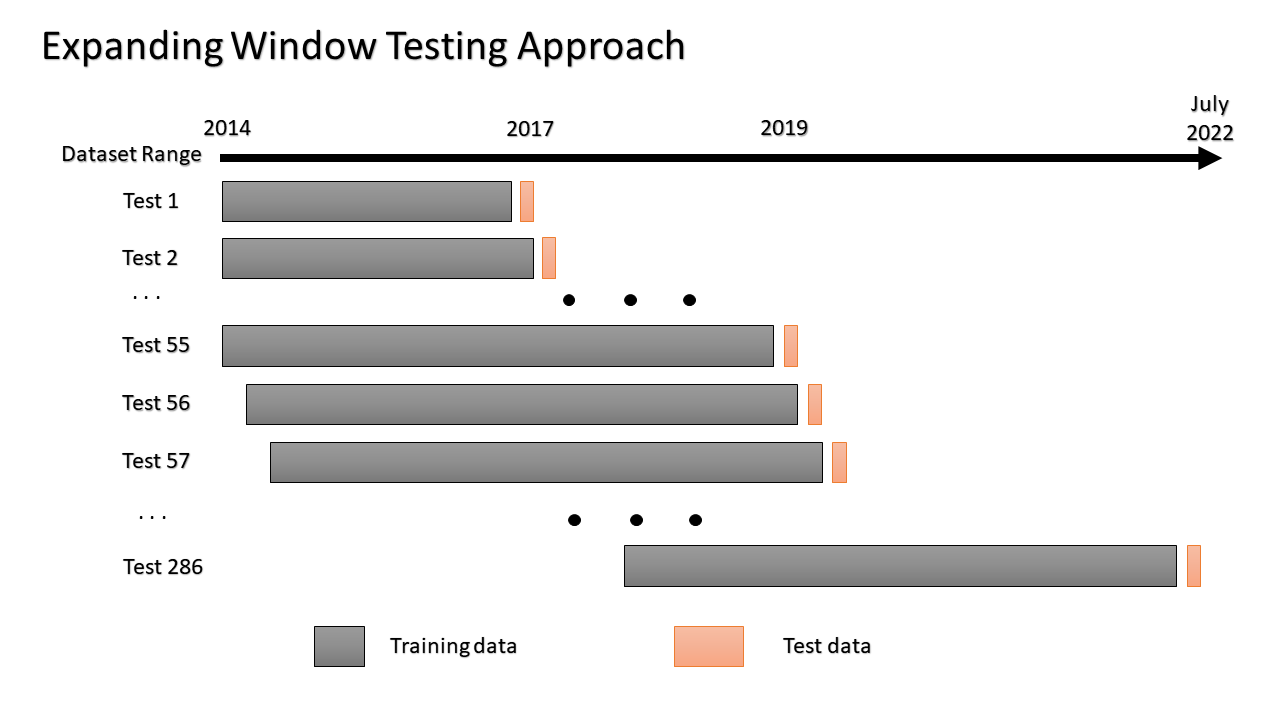}
\caption{The expanding window testing approach used in this study. Each testing window represents a one-week period. In total 286 training and testing cycles were performed for each model covering the periods between 2017 and July 2022.}
\label{figure:expwin}
\end{figure*}

The models were tested on data covering five and a half years from 2017 through to mid-2022. A modified version of the expanding window approach was used for estimating the generalisability of the models. Up to a maximum of 5 previous years of training data were used for creating each model\footnote{The models were trained on a maximum of five years of historic data in part due to the preference of more recent data for modelling, and due to a limitation in the acquisition of the Google Trends search terms data. Google Trends data can only be extracted for up to five years at a weekly granularity and beyond this, the data is aggregated at a monthly granularity. Since Google Trends data is scaled and provided as a relative index, it made it unsuitable to concatenate different periods together.}. Following each training phase, the models were then tested on forecasting 7 days ahead. There were in total of 286 test sets in this hold-out approach.  The models were initially trained on data from 2014-2016\footnote{Data prior to 2014 could not be used for training due to the autoregressive values which created missing values in the initial few years.} and the forecasts were then made starting from 1 January 2017 up to 7 days ahead. Figure  \ref{figure:expwin} visually depicts our training/testing methodology. The whole testing process was performed three times; once for models containing all the proposed features. The second time for models using only autoregressive features to establish the efficacy of the proxy features. Finally, the models were trained using only proxy features in order to both assess their utility and extract additional insights.  

\subsection{Model error measurements}  The models were evaluated using several metrics with each one providing a slightly different perspective.

Mean Absolute Percentage Error (MAPE)  is frequently used in literature and is recommended as the primary evaluation metric for forecasts \cite{bowerman2005forecasting}. We therefore followed this recommendation. MAPE is scale-independent and can be used to compare forecasts across datasets and studies with different ranges of values for the dependent variable. The calculation of MAPE is as follows:

\begin{equation}\label{MAPE}
   {\rm MAPE} = \frac{100}{T} \sum_{t=1}^T  
   \abs{\frac{\hat{y}_{t} - y_{t}}
   {y_t}  }
\end{equation}

\noindent where \textit{T} is the number of forecasts under evaluation and $\hat{y}_{t} - y_{t}$ is the error or residual term arising from the difference between the observed \textit{y} value and the forecast value $\hat{y}_{t}$ at time point \textit{t}. MAPE is a useful measure because it is able to express deviations between the observation and the forecasted values in terms of percentages, and as such, it is easy to interpret. 

We also note the Root Mean Square Error (RMSE) for each model as defined below:  

\begin{equation}\label{RMSE}
    {\rm RMSE} = \sqrt{\frac{1}{T} \sum_{t=1}^T (\hat{y}_{t} - y_{t})^2}
\end{equation}

RMSE is instructive since it describes the dispersion of the errors while being scaled to the dependent variable, therefore smaller RMSE values are preferred. 

For completeness, we also report the Mean Absolute Error for all models, which in conjunction with the previous two metrics is also used by some more recent studies in this field \cite{zhang2022forecasting}. MAE is the average absolute difference between the observation and the forecasted values. Given this property, it is to some degree conceptually easier to interpret and due to the squaring of the differences, it is less sensitive to large errors, unlike RMSE. Therefore, several significant errors will influence RMSE to a larger extent than MAE. The calculation for MAE is:
\begin{equation}\label{MAE}
    {\rm MAE} = \frac{1}{T} \sum_{t=1}^T   \abs{\hat{y}_{t} - y_{t}}
\end{equation}

Lastly, we extensively use mean ranks in order to concisely summarise the performance of all the algorithms across every test dataset. For each forecast period of 7-days, every algorithm was ranked from 1 to 15 with respect to its MAPE value, with the best performing algorithm achieving the rank of 1. This was performed across all 52 testing periods per year and across each of the 5.5 years of testing, and from this, the mean ranking was calculated.

\subsection{Statistical measurements}
   
In addition to the evaluation metrics, we also use Theil’s \textit{U} statistic \cite{theil1961economic} in order to assess the model accuracy relative to the persistence model where the forecast value is equal to the previous value. Since the method squares the errors, it gives more weight to large deviations and exaggerated them, which can serve as a useful method for identifying sub-optimal models. The Theil’s \textit{U} statistic value is calculated below as:

    \begin{equation}
   U = \sqrt{
        \frac{
            \frac{1}{T}
            \sum_{t=1}^{T-1} \left(\frac{\hat{y}_{t+1} - y_{t+1}}{y_t}\right)^2}
        {    
            \frac{1}{T} \sum_{t=1}^{T-1} 
               \left(\frac{y_{t+1} - y_t}{y_t}\right)^2
        }
      }
   \end{equation}
  
\noindent where \textit{y} again is the observed value and \textit{$\hat{y}_t$} is the forecast value at a given time step \textit{t}. When interpreting this statistic, values lower than 1 indicate that the model is performing better than the persistence model, while values of 1 and beyond indicate that the forecast accuracy is equivalent to the persistence model and is in fact worse as the values increase.

Finally, we report the \citet{Diebold1995} statistical test to establish whether the sequences of forecasts of the models are meaningfully different from those of the benchmark.  
In order to determine this, we use this test to compare the outputs of the competing model with those of the benchmark estimation models, with respect to the observed values. 

\subsection{Model interpretability and explainability} 

The emerging field of XAI, spurred on by increasing regulatory requirements \cite{mathrani2021perspectives}, addresses the challenge posed by uninterpretable models and attempts to answer both the "how" and the "why" of their decision-making in approximate terms.
A set of approaches called post-modelling explainability tools, aim to answer "how" an algorithm behaves in the construction of a model during the training process, resulting in model interpretability, as well as "why" the generated model has made a specific prediction/forecast, resulting in model explainability \cite{minh2021explainable}.  
One of the techniques that currently stands out as state-of-the-art for extracting the interpretability and explainability of predictive models \cite{gramegna2021shap}, is SHAP \cite{Lundberg2017}, which is used in this study.

We employ SHAP to examine the internal mechanics of the predictive models at both the \textit{global} and \textit{local} levels. At a global level of analysis, we seek to understand the overall effects that each feature exerts on the model outputs. We primarily use feature importance plots to gain this insight which typically ranks as well as depicts the relative impacts of each feature. We also examine the effects that changing feature values have on the final forecast. Additionally we use feature dependence plots in order to shed light on how pairs of features interact in order to affect the final forecast. These plots together offer a degree of  \textit{high-level} interpretability of the main drivers for a given model. In considering model behaviour at a local level, we attempt to extract a model's reasoning as to precisely why exactly it has produced a given forecast for a specific data point. 

SHAP generates new models which approximate the forecasting behaviour of the underlying "black-box" models. These models are called \textit{surrogate models} and are designed to be more interpretable. 
 
SHAP (an abbreviation for SHapley Additive exPlanation) itself is based on Shapley values \cite{shapley1953quota}. The technique operates on the principles of game theory. It attributes each feature's marginal contribution to the final predictive outcome in collaboration with the other features. In this way, SHAP is able to provide both \textit{global} interpretability and \textit{local} explainability.

\section{Results}\label{sec2}

The results are presented in two parts. The first examines the accuracies and the statistical significance of the forecasting models, together with the efficacy of the proxy features. The effects of the concept drift on the model accuracies are highlighted from the year 2020 onward. The second part analyses both the interpretability and the explainability aspects of the models, with a focus on determining the utility of the proxy variables. The second part also attempts to extract insights concerning the model behaviour and the underlying features.

\subsection{Forecast Modelling}\label{fm}

Table \ref{table:Average_Metrics} shows a high-level summary of all the models across both clinics, displaying the MAPE values for each candidate model developed with a full set of features. To establish the utility of the proxy features, each model's MAPE score is contrasted with the MAPE values of models developed using only autoregressive features, which are placed adjacently in parentheses. Both the RMSE and MAE values are also listed, together with the average rank scores based on MAPE. The table indicates that across both clinics the proposed models have outperformed those that have not used proxy features. At the bottom of the table, it can be seen that the best-performing algorithm was Voting which combined machine learning and standard statistical approaches.

\begin{table*}[hbt]
\fontsize{7pt}{9pt} 
\selectfont
\caption{Forecast accuracies by algorithm and clinic using all features as well as MAPE accuracies of models using only autoregressive features in parentheses. Ranks per clinic are based on MAPE and are also combined across both.  } 
\centering
\setlength\tabcolsep{3pt}
\begin{tabular}{lccccccccc}

\hline
 & \multicolumn{4}{c}{Clinic 1} 
 & \multicolumn{4}{c}{Clinic 2} 
 & \multicolumn{1}{c}{Combined} 
 \\
 & MAPE  & Rank & RMSE & MAE 
& MAPE  & Rank & RMSE & MAE & Rank 
\\
\hline
kNN & 17.5$\pm$8.8 (11.3) & 12.7 & 29.8 & 25.2 & 21.7$\pm$14.7 (15.0)  & 12.3 & 15.7 & 13.2 & 12.5 \\   SVR (NU) & 17.7$\pm$11.3 (11.2) & 12.4 & 29.7 & 25.3 & 20.4$\pm$16.8 (14.5) & 11.1 & 14.6  & 12.4 & 11.7 \\  \textbf{Benchmark} & 16.7$\pm$13.5 (16.7) & 11.4 & 25.6 & 22.0 & 21.9$\pm$21.1 (21.9) & 11.1 & 14.6  & 12.4 & 11.2 \\   Naive & 16.3$\pm$11.7 (16.3) & 11.0 & 25.7 & 22.3 & 19.1$\pm$16.6 (19.1) & 10.3 & 13.5  & 11.4 & 10.6 \\   Prophet & 10.1$\pm$5.3 (11.8) & 6.9 & 16.7 & 14.2 & 15.7$\pm$14.6 (16.6) & 7.9 & 10.9 & 9.2  & 7.4 \\   Naive (Enhanced) & 10.7$\pm$6.1 (10.7) & 7.6 & 17.7 & 14.9 & 14.1$\pm$9.4 (14.1) & 7.0  & 10.1 & 8.4 & 7.3 \\   ARIMA & 10.5$\pm$5.9 (10.5) & 7.4 & 17.3 & 14.6 & 14.4$\pm$9.6 (14.1) & 7.2 & 10.1 & 8.5  & 7.3 \\   Kernel Ridge & 9.9$\pm$4.2 (10.6) & 7.0 & 16.8 & 14.1 & 14.1$\pm$9.0 (14.3) & 7.5 & 10.1  & 8.3 & 7.3 \\   Ridge & 9.9$\pm$4.2 (10.6) & 6.9 & 16.8 & 14.1 & 14.1$\pm$9.2 (14.3) & 7.3 & 10.1 & 8.3 & 7.1 \\   Random Forest & 10.0$\pm$5.6 (10.5) & 7.1 & 16.7 & 13.9 & 13.9$\pm$9.1 (14.2) & 7.1  & 10.0 & 8.3 & 7.1 \\   CatBoost & 9.6$\pm$4.4 (10.4) & 6.8 & 16.3 & 13.5 & 13.4$\pm$7.0 (14.2) & 7.1 & 9.8 & 8.1  & 7.0 \\   Gradient Boosting & 9.6$\pm$4.5 (10.3) & 6.6 & 16.3 & 13.5 & 13.6$\pm$7.6 (14.1) & 7.1  & 9.9 & 8.2 & 6.8 \\   Stacking & 9.4$\pm$4.8 (10.0) & 5.8 & 15.8 & 13.2 & 13.6$\pm$10.4 (13.9) & 6.2 & 9.6 & 8.0  & 6.0 \\   Averaging & 9.2$\pm$4.6 (9.9) & 5.1 & 15.3 & 12.8 & 13.3$\pm$9.4 (13.7) & 5.6  & 9.5 & 7.9 & 5.4 \\   Voting & 9.0$\pm$4.6 (9.7) & 5.0 & 15.0 & 12.6 & 13.1$\pm$9.1 (13.5) & 5.4 & 9.3 & 7.8 & 5.2
\\
\hline
\end{tabular}
\label{table:Average_Metrics}
\end{table*}

The table combines the rank-ordering of the modelling techniques from worst performing to most accurate with respect to MAPE across both clinics (seen in the final column of Table \ref{table:Average_Metrics}). The results indicate that the Voting algorithm has consistently outperformed all other techniques on this dataset by leveraging the advantages of both machine learning and statistical approaches. The top three performing techniques are all ensemble-based meta approaches which have in various ways combined the models from the underlying algorithms. These approaches were closely followed by standard ensemble-based methods, with Gradient Boosting-based approaches outperforming Random Forest. The rank-ordered results also indicate that the existing in-house benchmark approach to patient forecasting has been significantly improved upon by the best-performing methods. 

It is also observable through MAPE scores that the predictability of Clinic 2 patient flows is generally lower than that of Clinic 1. This is attributable to overall lower patient flows at Clinic 2 which predisposes it to more variability and thus higher unpredictability. This result underscores the limitations of blindly using the MAPE measure for comparisons across different studies without taking volumes into consideration since the magnitude of the total patient volumes affects variability and consequently, predictability. In saying that, comparative studies from reported literature cite MAPE accuracies ranging from 7.6\% to 10.7\% for 7-day forecasts (Table \ref{table:litsummary}) which provides some context for the 9\% and 13.1\% accuracies achieved by Clinics 1 and 2 respectively, taking into account that the prior studies did not cover the pandemic period.

\begin{figure}[hbt]
\centering
\includegraphics[scale=0.3]{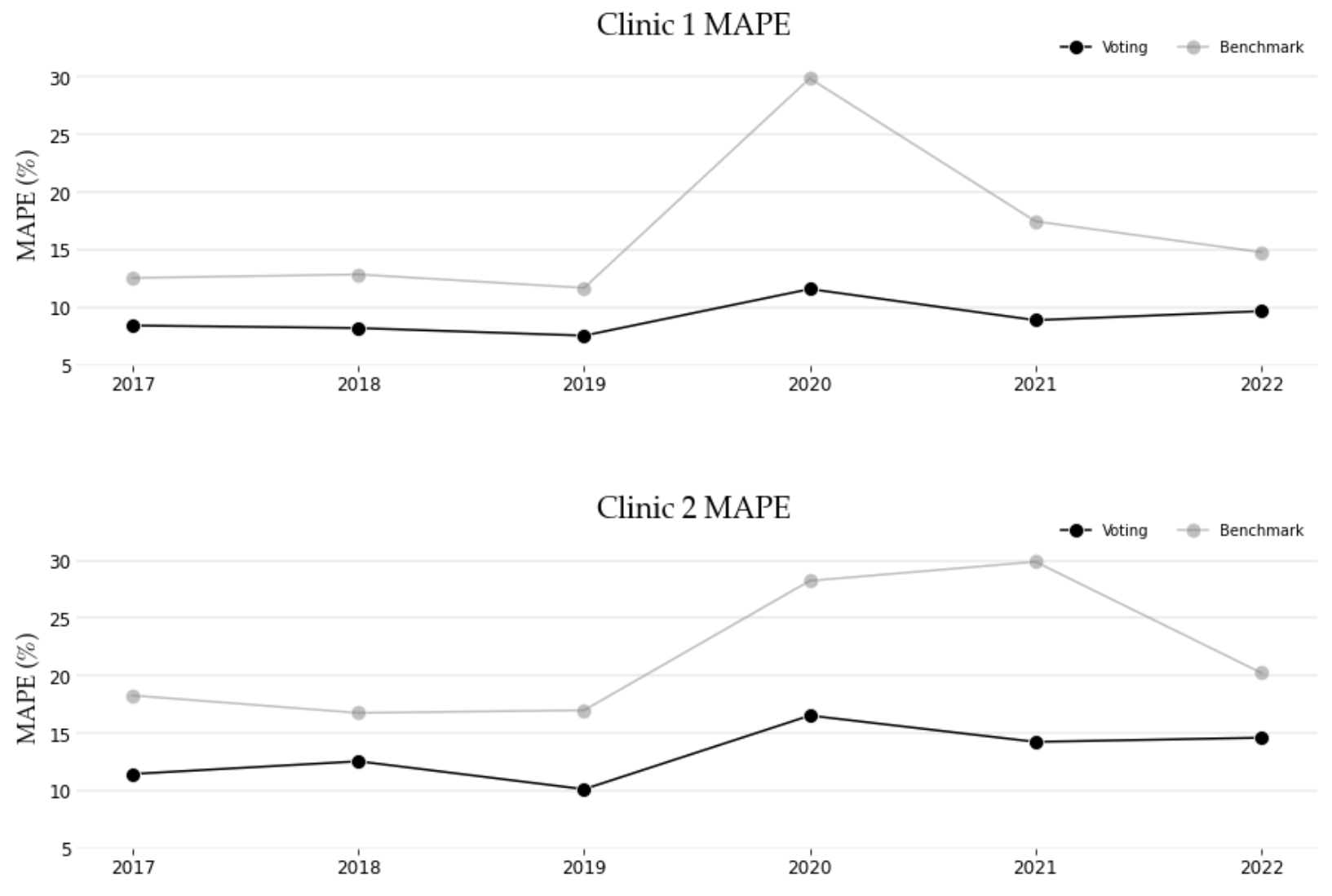}
\caption{Comparison of MAPE accuracies between the benchmark and Voting models across both clinics and all the years.}
\label{figure:mapes}
\end{figure}

To highlight the effects of the pandemic-induced concept drift, Tables \ref{table:Metrics_S} and \ref{table:Metrics_N} examine the accuracies of the models by each year, across both clinics. Firstly, the tables confirm that the Voting method has consistently generated higher generalisability across all years according to the combined mean rank. However, an abrupt pandemic-induced deterioration in the accuracies can be seen for 2020 onward on both tables. The accuracy declined by 55\% and 63\% for the Voting method across Clinic 1 and 2 respectively from 2019 to 2020, before beginning to improve from 2021 onward. The subsequent improvement in the accuracies after 2020 indicates that the models made some measure of adjustment to the concept drift and have been able to adapt to the new conditions with the help of the proxy features. In comparison, the benchmark forecasts deteriorated by 156\% and 66\% across Clinic 1 and 2 respectively for the same period. To visually highlight the effects of the concept drift on the accuracies and the subsequent adaptations by the models, we render the MAPE values for both clinics, contrasting Voting and benchmark models in Figure \ref{figure:mapes}. Sharp deteriorations especially for the benchmark model can be seen in 2020 for both clinics, together with adjustments which occur faster for the Voting model.

\begin{table*}[hbt]
\caption{Accuracies and mean ranks for all models across each year for Clinic 1.} 
\fontsize{7pt}{9pt} 
\selectfont
\centering
\setlength\tabcolsep{2.5pt}
\begin{tabular}{lccccccccccccc}
\hline
 & \multicolumn{2}{c}{2017} 
 & \multicolumn{2}{c}{2018} 
 & \multicolumn{2}{c}{2019} 
 & \multicolumn{2}{c}{2020} 
 & \multicolumn{2}{c}{2021} 
 & \multicolumn{2}{c}{2022} &
\\
& MAPE & R & MAPE & R & MAPE & R & MAPE & R 
& MAPE & R & MAPE & R & Mean R
\\
\hline
kNN & 16.8 & 13.2 & 16.1 & 13.3 & 15.4 & 13.4 & 22.7 & 11.8 & 16.1 & 12.2  & 17.8 & 12.3 & 12.7 \\   SVR (NU) & 17.2 & 13.8 & 15.5 & 12.9 & 14.9 & 13.2 & 27.2 & 11.6 & 15.2 & 10.9 & 15.2 & 11.5  & 12.4 \\   \textbf{Benchmark} & 12.6 & 10.3 & 12.9 & 11.3 & 11.7 & 11.8 & 29.9 & 12.4 & 17.5 & 11.4 & 14.8 & 10.8  & 11.4 \\   Naive & 13.4 & 11.1 & 12.6 & 10.9 & 12.1 & 11.5 & 25.6 & 11.0 & 18.7 & 11.1 & 14.1 & 9.9 & 11.0 \\   Naive (Enhanced) & 10.0 & 8.2 & 9.3 & 7.5 & 8.6 & 7.3 & 14.9 & 8.0 & 11.1 & 7.7 & 9.7 & 6.3 & 7.6 \\   ARIMA & 9.7 & 7.0 & 9.7 & 8.7 & 9.0 & 7.7 & 13.1 & 7.2 & 10.8 & 6.7 & 10.5 & 6.8 & 7.4 \\   Random Forest & 9.3 & 7.7 & 9.0 & 7.0 & 8.1 & 6.9 & 13.9 & 7.3 & 9.9 & 7.0 & 9.8 & 6.0 & 7.1 \\   Kernel Ridge & 9.0 & 6.2 & 9.2 & 7.1 & 7.9 & 5.9 & 12.5 & 6.9 & 10.6 & 8.2 & 10.7 & 8.8 & 7.0 \\   Ridge & 9.0 & 6.2 & 9.2 & 7.2 & 7.9 & 6.0 & 12.4 & 6.6 & 10.4 & 7.7 & 10.8 & 8.9 & 6.9 \\   Prophet & 8.4 & 4.9 & 8.4 & 5.8 & 8.2 & 7.1 & 13.5 & 8.1 & 10.2 & 7.3 & 14.0 & 9.8 & 6.9 \\   CatBoost & 9.7 & 8.7 & 8.7 & 6.3 & 8.1 & 7.0 & 11.9 & 5.8 & 9.7 & 6.8 & 9.7 & 6.0 & 6.8 \\   Gradient Boosting & 9.2 & 7.5 & 8.9 & 6.5 & 8.0 & 6.6 & 12.2 & 6.1 & 9.8 & 6.5 & 9.7 & 6.3  & 6.6 \\   Stacking & 8.4 & 4.9 & 8.3 & 5.2 & 7.8 & 5.8 & 12.2 & 6.5 & 9.7 & 6.6 & 11.1 & 6.4 & 5.8 \\   Averaging & 8.5 & 5.5 & 8.2 & 4.8 & 7.6 & 5.0 & 11.9 & 5.4 & 9.2 & 5.0 & 9.9 & 5.3 & 5.1 \\   Voting & 8.4 & 4.8 & 8.2 & 5.4 & 7.5 & 4.8 & 11.6 & 5.3 & 8.9 & 4.8 & 9.7 & 4.8 & 5.0\\
\hline
\end{tabular}
\label{table:Metrics_S}
\end{table*}

\begin{table*}[hbt]
\caption{Accuracies and mean ranks for all models across each year for Clinic 2. }
\fontsize{7pt}{9pt} 
\selectfont
\centering
\setlength\tabcolsep{2.5pt}
\begin{tabular}{lccccccccccccc}
\hline
 & \multicolumn{2}{c}{2017} 
 & \multicolumn{2}{c}{2018} 
 & \multicolumn{2}{c}{2019} 
 & \multicolumn{2}{c}{2020} 
 & \multicolumn{2}{c}{2021} 
 & \multicolumn{2}{c}{2022} &
\\
& MAPE & R & MAPE & R & MAPE & R & MAPE & R 
& MAPE & R & MAPE & R & Mean R
\\
\hline
kNN & 17.4 & 12.2 & 20.4 & 13.0 & 19.5 & 12.8 & 27.9 & 12.2 & 24.1 & 11.5  & 20.2 & 11.9 & 12.3 \\   \textbf{Benchmark}  & 18.3 & 11.7 & 16.8 & 10.3 & 17.0 & 11.4 & 28.2 & 9.5 & 29.9 & 12.7 & 20.2 & 10.9  & 11.1 \\   SVR (NU) & 15.4 & 11.0 & 19.1 & 11.8 & 18.9 & 12.7 & 28.4 & 10.7 & 22.1 & 9.8 & 16.9 & 9.6 & 11.0 \\   Naive & 16.4 & 10.4 & 17.3 & 10.2 & 17.9 & 11.8 & 23.6 & 7.6 & 20.8 & 11.5 & 18.9 & 10.0 & 10.3 \\   Prophet & 12.4 & 6.5 & 13.1 & 6.9 & 11.2 & 6.5 & 23.8 & 10.0 & 16.9 & 8.2 & 18.3 & 10.2 & 7.9 \\   Kernel Ridge & 13.2 & 8.5 & 13.7 & 7.9 & 11.2 & 7.3 & 17.4 & 7.6 & 14.9 & 7.0 & 13.9 & 6.2  & 7.5 \\   Ridge & 13.0 & 7.8 & 13.7 & 7.8 & 11.3 & 7.3 & 17.5 & 7.5 & 14.9 & 6.7 & 13.8 & 6.0 & 7.3 \\   ARIMA & 12.3 & 6.3 & 13.8 & 7.4 & 11.0 & 6.2 & 17.4 & 8.3 & 16.4 & 7.4 & 16.1 & 8.0 & 7.2 \\   CatBoost & 12.6 & 7.7 & 13.3 & 7.4 & 11.2 & 7.2 & 14.6 & 6.2 & 14.2 & 6.7 & 15.6 & 7.9 & 7.1 \\   Gradient Boosting & 12.7 & 7.9 & 13.4 & 6.8 & 11.2 & 7.0 & 15.7 & 7.2 & 14.8 & 6.8 & 14.7  & 6.6 & 7.1 \\   Random Forest & 12.4 & 7.4 & 13.4 & 6.9 & 11.3 & 7.4 & 16.5 & 6.6 & 15.2 & 6.9 & 15.1 & 7.2  & 7.1 \\   Naive (Enhanced) & 12.2 & 6.5 & 13.3 & 7.1 & 11.6 & 7.1 & 17.3 & 7.6 & 16.4 & 7.5 & 13.4 & 5.3  & 7.0 \\   Stacking & 12.0 & 6.2 & 12.6 & 5.7 & 10.3 & 5.2 & 17.9 & 6.8 & 14.5 & 6.3 & 15.5 & 7.6 & 6.2 \\   Averaging & 11.6 & 5.2 & 12.6 & 5.5 & 10.2 & 5.1 & 16.7 & 6.1 & 14.2 & 5.8 & 15.1  & 6.8 & 5.6 \\   Voting & 11.5 & 4.8 & 12.6 & 5.3 & 10.1 & 5.1 & 16.5 & 6.1 & 14.3 & 5.4 & 14.6 & 5.8 & 5.4
\\
\hline
\end{tabular}
\label{table:Metrics_N}
\end{table*}

Returning to the models' accuracies by year, the overall improvements of the best model (Voting) over the benchmark are $\thicksim$33\% and $\thicksim$23\% for Clinics 1 and 2 respectively. A detailed presentation of the percentage improvement achieved by the proposed models over the benchmark approach by year as well as overall can be seen in Tables \ref{table:Perc_Imp_S} and \ref{table:Perc_Imp_N} in Appendix \ref{appendix:improvements}. The TheilU statistic comparing the Voting model to the benchmark showed that values are below 1, indicating an improvement over the benchmark. A full summary of these values can be seen in Tables \ref{table:TheilU_S} and \ref{table:TheilU_N} in  Appendix \ref{appendix:improvements}. Finally, in establishing the advantage of the Voting model over the benchmark, the Diebold-Mariano test across both clinics and for each year indicated that all Voting model forecasts can be considered significantly different (at a 1\% level) to the benchmark approach.

\begin{figure}[hbt]
\centering
\includegraphics[scale=0.65]{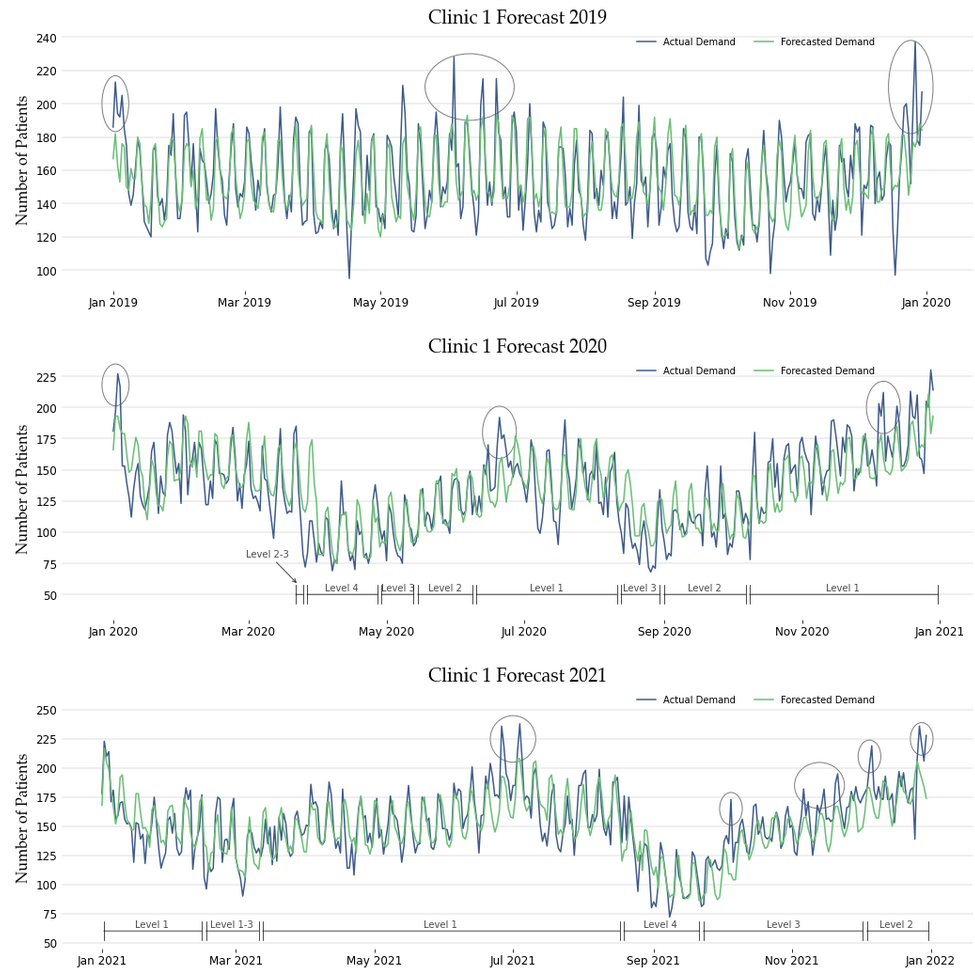}
\caption{Example of forecasts versus observations between 2019 and 2021 with the corresponding pandemic Alert Levels on the x-axis highlighted.}
\label{figure:forecasts}
\end{figure}

Next, we depict an example of the Voting model's forecasting behaviour across three distinctive years, namely from 2019 to 2021, which cover both the stable (2019) and the concept-drift periods (2020-2021) in Figure \ref{figure:forecasts}. This figure shows the forecasting results for estimating patient flows for Clinic 1 while contrasting them with the observed values. Relatively stable patient flows together with accurate forecasts can be seen for 2019, with several undetected patient surges occurring in that year (circled) where the discrepancy between the forecasts and the actuals was 20\% or above. Significant COVID-19-related disruptions and the ensuing concept drift can be seen in the figures depicting 2020 and 2021  when recurring lockdowns and GP closures took place. In the figures, the forecasts largely demonstrate a high degree of adaptability to the underlying changes aided by the proposed proxy features which are able to incorporate new information. While the forecasts were generally effective, points of interest have been highlighted in the figures where the observed values exceeded the foretasted values by more than 20\%.

To further establish the efficacy of the proxy features,  Table \ref{table:proxy_features} contrasts MAPE results of the benchmark, Voting model using all features and the Voting model using only proxy features for Clinic 1, across all years. Though the full-feature model was superior, the results indicate that the Voting model using proxy features only was able to account for a considerable amount of variation in the patient flows alone. During stable periods (2017-2019) the full-feature model improved over the proxy-only model between $\thicksim 18\%$ and $\thicksim 31\%$. However, during the concept drift period (2020-2022), this was reduced to only a $\thicksim 7\%$ to $\thicksim 16\%$ improvement. The experiment indicates that a high degree of information is entailed within the proxy variables and that they are able to explain a significant degree of variation in the dependent variable, especially during the particular case of the recent pandemic-induced concept drift.

\begin{table*}[hbt]
\caption{Contrasting MAPE accuracies between models using all features versus those generated by proxy-only features for Clinic 1. } 
\centering
\setlength\tabcolsep{8pt}
\begin{tabular}{lcccccc}
\hline
 & \multicolumn{1}{c}{2017} 
 & \multicolumn{1}{c}{2018} 
 & \multicolumn{1}{c}{2019} 
 & \multicolumn{1}{c}{2020} 
 & \multicolumn{1}{c}{2021} 
 & \multicolumn{1}{c}{2022} 
\\
\hline
\textbf{Benchmark}  & 12.6  & 12.9  & 11.7 &  29.9  & 17.5 & 14.8  \\
Voting  (proxy-features) & 11.0  & 9.7  & 9.7 & 12.4  & 10.0  & 11.3   \\
Voting (all features) & 8.4  & 8.2  & 7.5  & 11.6  & 8.9  & 9.7\\

\hline
\end{tabular}

\label{table:proxy_features}
\end{table*}

\subsection{Model Interpretability and Explainability}\label{mie}

\begin{figure}[hbt]
\begin{tabular}{cc}

\subfloat(a){%
\includegraphics[trim=0cm 0.2cm 0cm 0cm, clip=true,scale=0.3]
{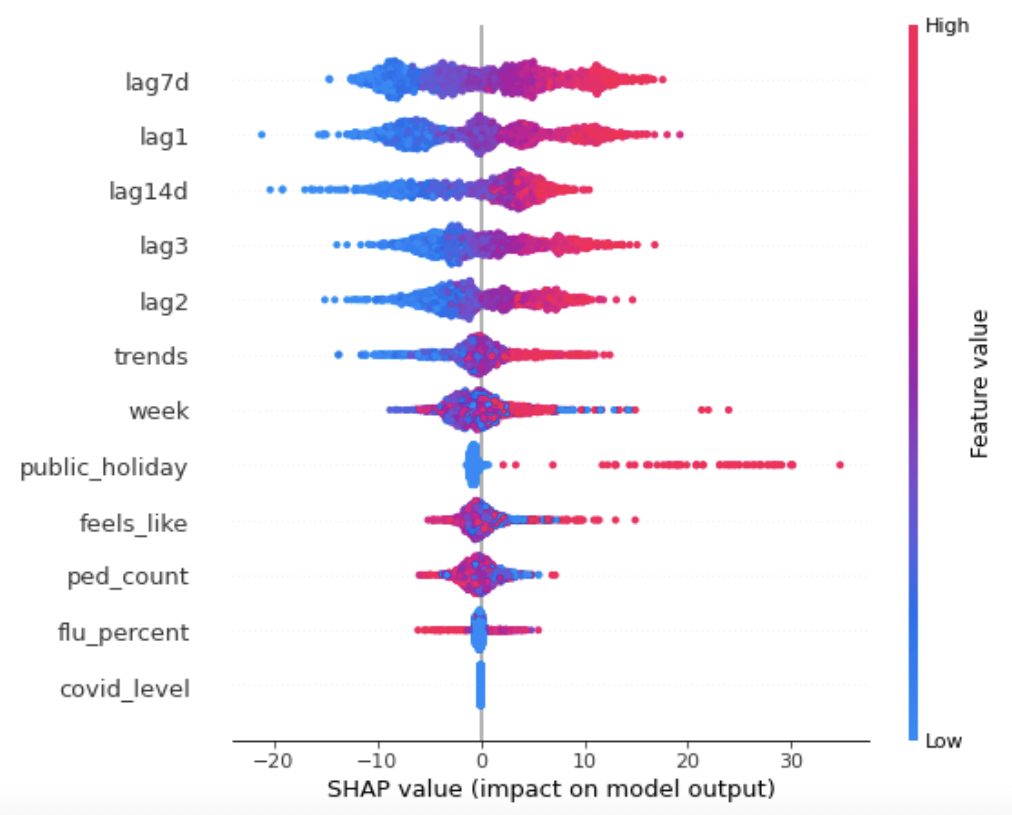}} & 

\subfloat(b){%
\includegraphics[trim=0cm 0.2cm 0cm 0cm, clip=true,scale=0.3]{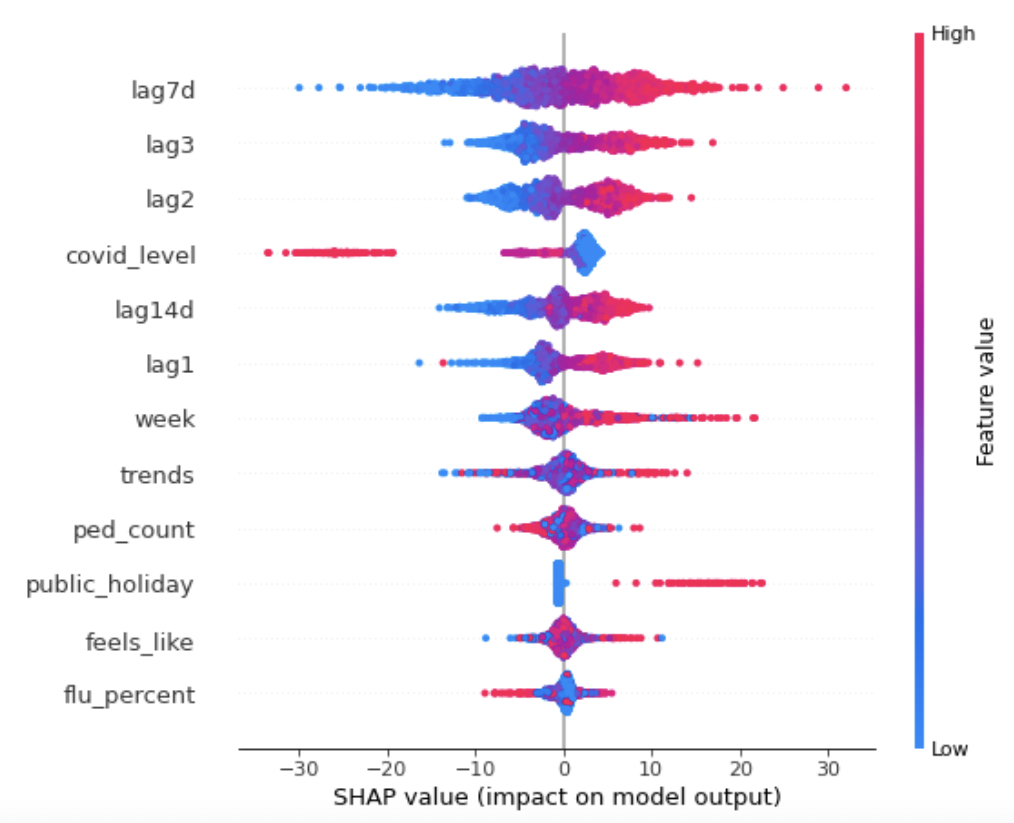}
} \\
\subfloat(c){%
\includegraphics[
scale=0.2]{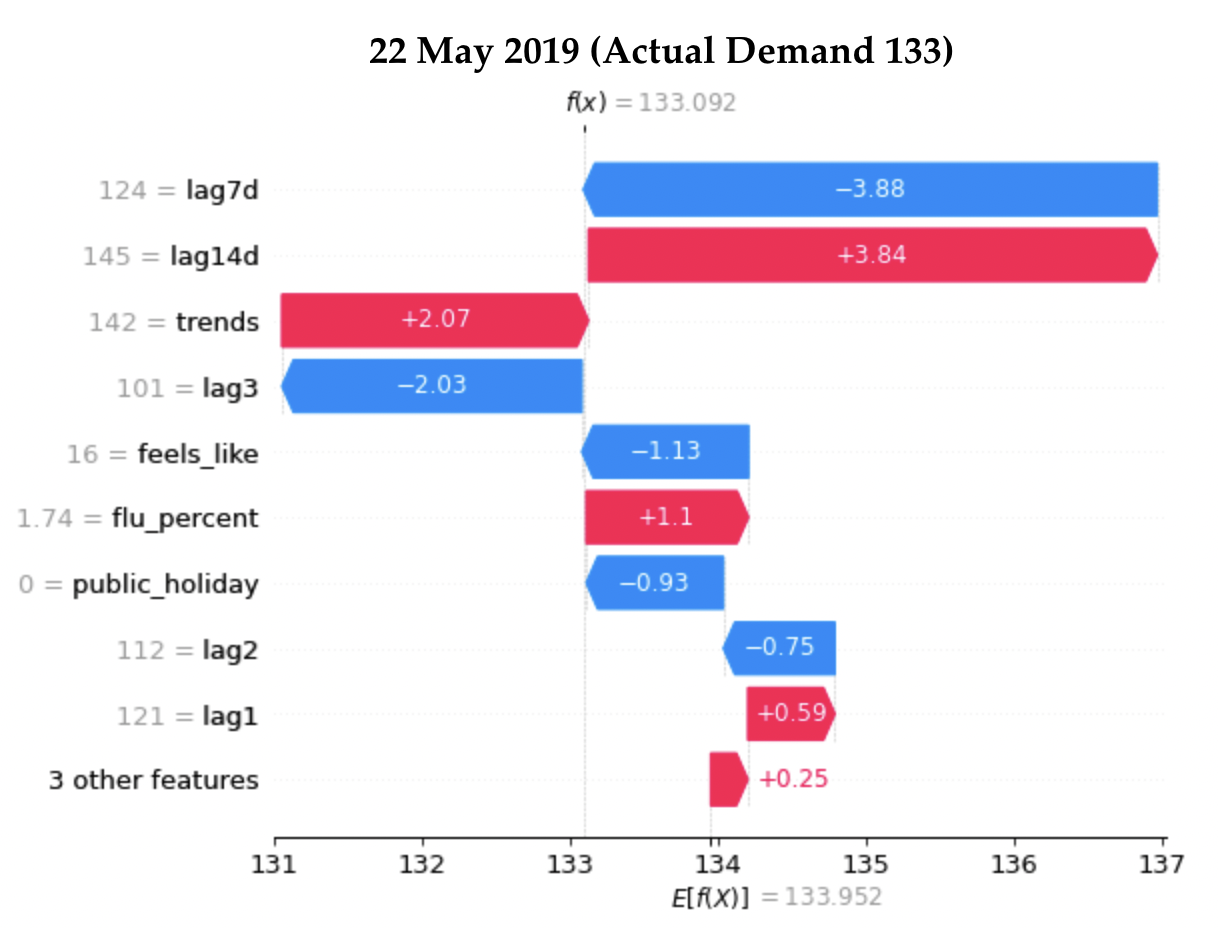}} & 
\subfloat(d){%
\includegraphics[
scale=0.2]{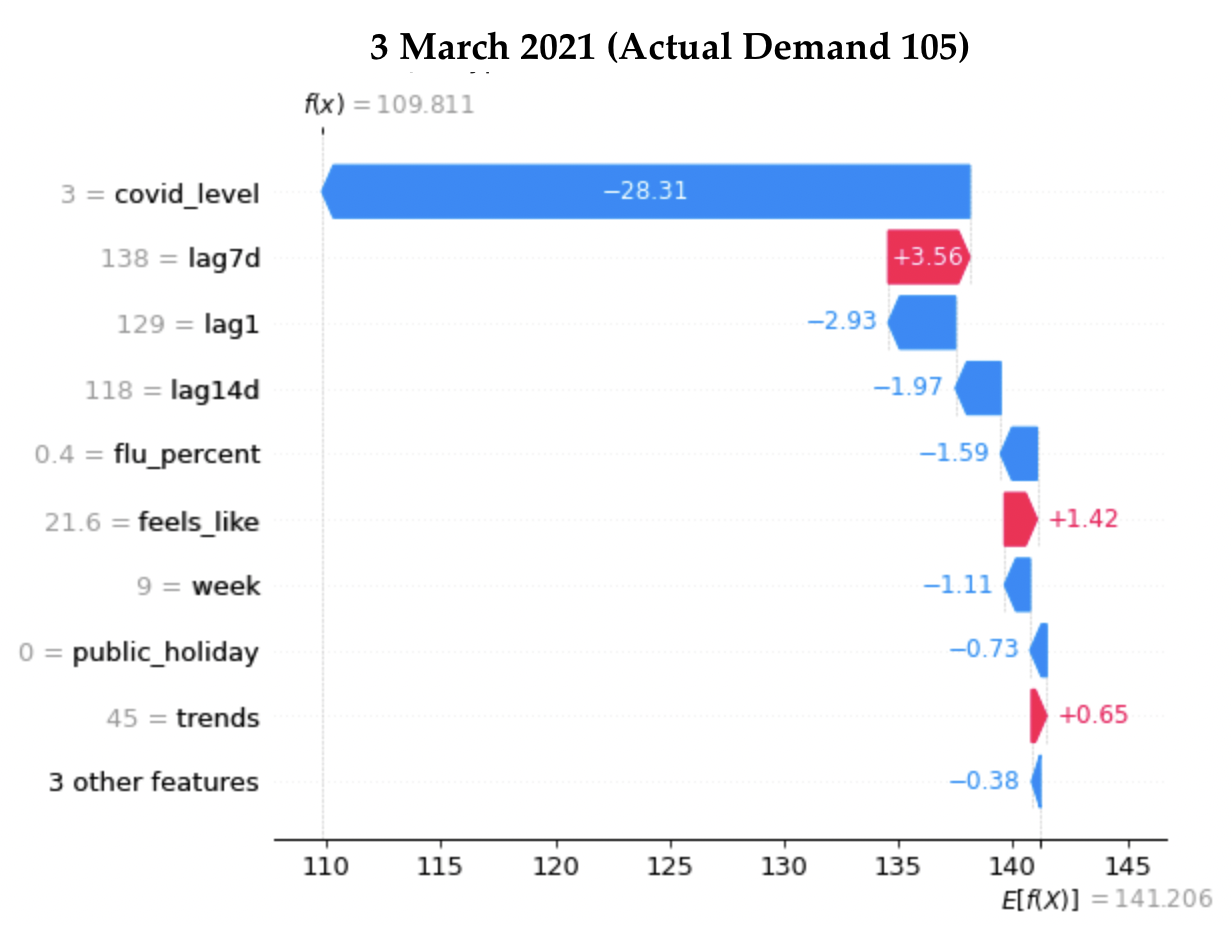}
} \\
\subfloat(e) {%
\includegraphics[
scale=0.2]{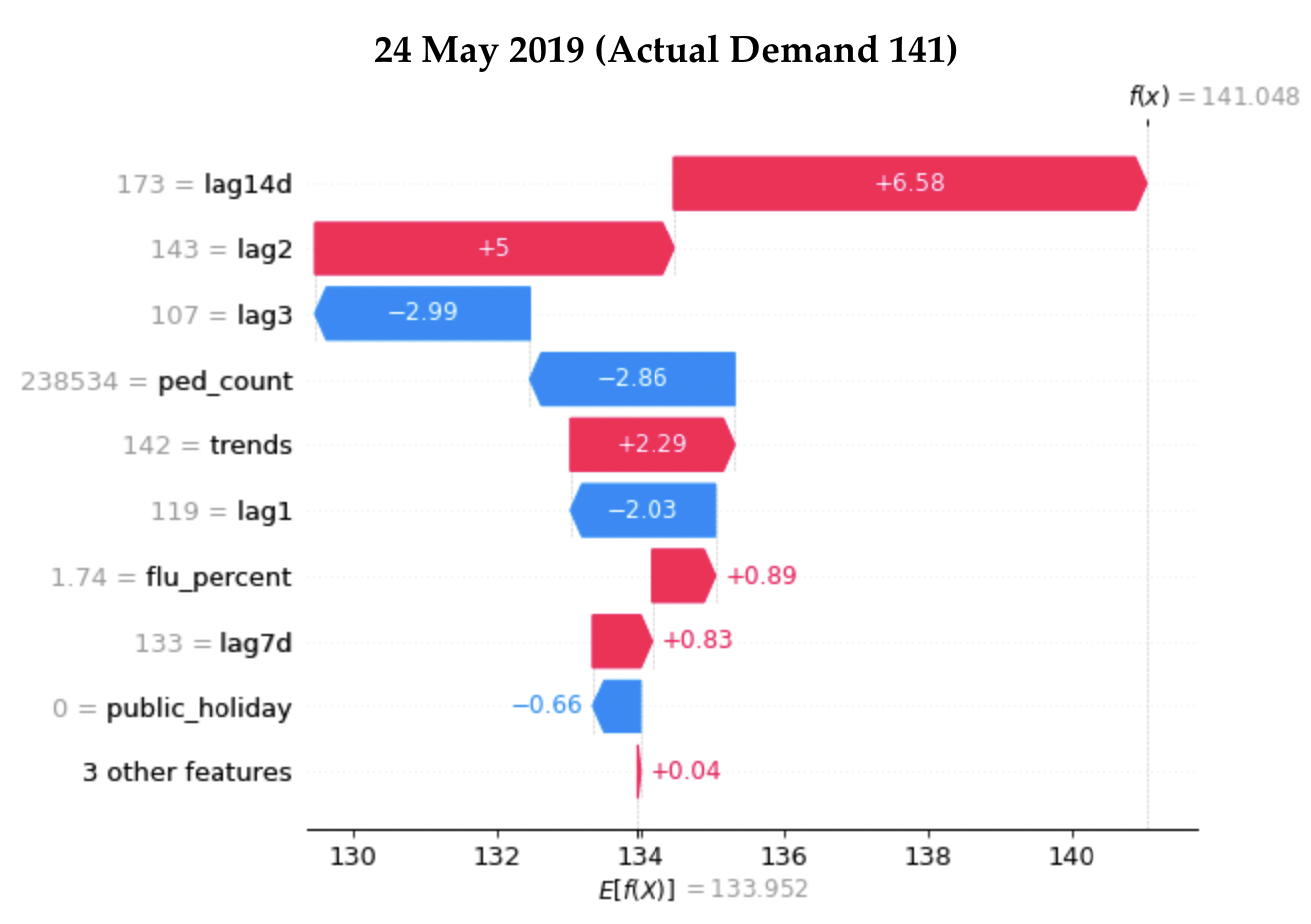}} & 
\subfloat(f){%
\includegraphics[
scale=0.2]{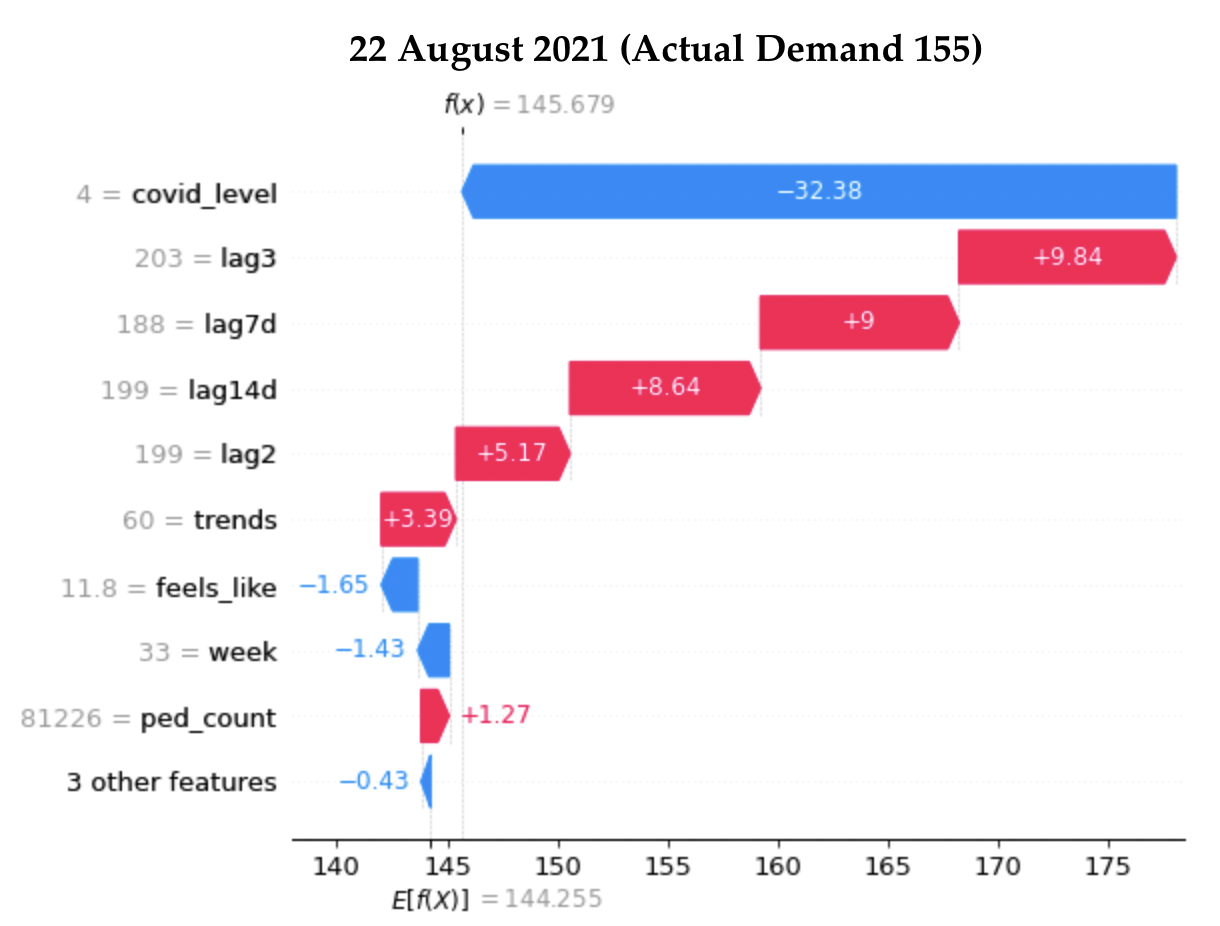}
} \\

\end{tabular}
\caption{SHAP graphs for 2019 (Left) and 2021 (Right) models. (a) and (b) depict feature importance summary plots. (c) - (f) show explanations of forecasts for specific dates together with the observed values.}
\label{Fig:Shap_all_features}
\end{figure}

We now use the SHAP technique in order to extract the Voting model's interpretability and the explainability of its forecasts. Figure \ref{Fig:Shap_all_features}a and  \ref{Fig:Shap_all_features}b depict the high-level interpretability of the model behaviour up to the end of 2019 and 2021 respectively. Again, the two years are selected to highlight the response of the models to the concept drift due to prevailing pandemic conditions affecting patient flows. The figures show the feature importance plots as determined by the models from most to least impactful, while also depicting the relative magnitude of the effect that each feature exerts on the final forecast. The figures also communicate how changes in the values of each feature drive the model's forecast upwards or downwards.

The two figures are in agreement that the most important feature influencing the forecast of patient demand up to the year they represent, are values from seven days prior, while values from two weeks prior and years are also prominent. However, it is clear that the prevailing COVID-19 Alert Level has gained a high-ranking position for importance in 2021. Generally, the importance of autoregressive features has been interpreted as more important by the models than that of the other proxy variables.

The SHAP summary plots in Figures \ref{Fig:Shap_all_features}a and \ref{Fig:Shap_all_features}b  communicate an additional dimension concerning global interpretability. In these figures, one can observe how an increase/decrease in the values of each underlying feature impacts the final forecast. The colours represent feature values where red equates to high values and blue to low values. The x-axis conveys the range of impact on the final forecast. The data points which have a positive SHAP value and appear to the right of the vertical zero line, have an impact on the forecast value which drives it towards predicting higher patient flow numbers. Conversely, the data points that have a negative SHAP value, to the left of the vertical zero line, decrease the forecasted patient flow numbers. As the points extend further from the zero vertical line, their effect and contribution to the final forecast correspondingly increase.

From Figures \ref{Fig:Shap_all_features}a and \ref{Fig:Shap_all_features}b, a general pattern can be seen where the higher values of the features have a correlated positive effect on pushing the forecasts toward higher totals. In Figure \ref{Fig:Shap_all_features}b, an exception can be seen in the effects of the COVID-19 Alert Level, where higher levels result in decreasing the forecasted patient flow. The impact of this feature as well as that of the public holidays is also asymmetrical. In other words, the model responds more strongly to forecasting lower patient demand if the values for the COVID-19 Alert Level are high, then it would predict a higher patient flow if the values for these features were low. The reverse holds for the public holiday feature. Some ambiguity exists in the effects that the values of 'feels like', 'pedestrian count' as well as 'flu percent' have on the final prediction.

Figures \ref{Fig:Shap_all_features}c-f expose the explainability of the models for a selection of specific forecast dates. These figures depict the top nine features for each forecast and their values on the y-axis. The features are rank-ordered by their impact on the final prediction. The figure can be interpreted as a contest of forcing effects between all the features. The expected value of each figure is denoted as $E[f(x)]$ which is the average value of all the forecasted data points. A final SHAP value at the top represents the eventual forecasted outcome. Blue bars represent the forcing effects towards smaller forecast patient flows, while red indicates the opposite. The size of the bars represents the effect size that each feature and its corresponding values exert. These graphs are best interpreted from the bottom up.

The forecasts for Figure \ref{Fig:Shap_all_features}c and \ref{Fig:Shap_all_features}e were made in 2019 which explains why the COVID-19 Alert Level did not play a role in the forecasts. In Figure \ref{Fig:Shap_all_features}c, the 7 and 14-day autoregressive features were the most impactful; however, their effects cancelled each other out, while Google search terms pushed the forecasts towards higher estimates. In contrast, Figure \ref{Fig:Shap_all_features}e shows that the autoregressive value from two years before had a strong effect on higher forecasts, and pedestrian counts featured more prominently in pushing the final forecast down. Figures \ref{Fig:Shap_all_features}d and \ref{Fig:Shap_all_features}f represent the concept-drift phase. In these figures the strong influence of the COVID-19 Alert Level can be seen responding to the changes in the dependent variable, pushing both forecasts down due to the high prevailing pandemic situation. In both cases, the pandemic level eclipsed the remaining features and successfully rectified the final forecast toward an accurate estimation.  


\begin{figure}[hbt]
\begin{tabular}{cc}
\subfloat(a){%
\includegraphics[scale=0.3]{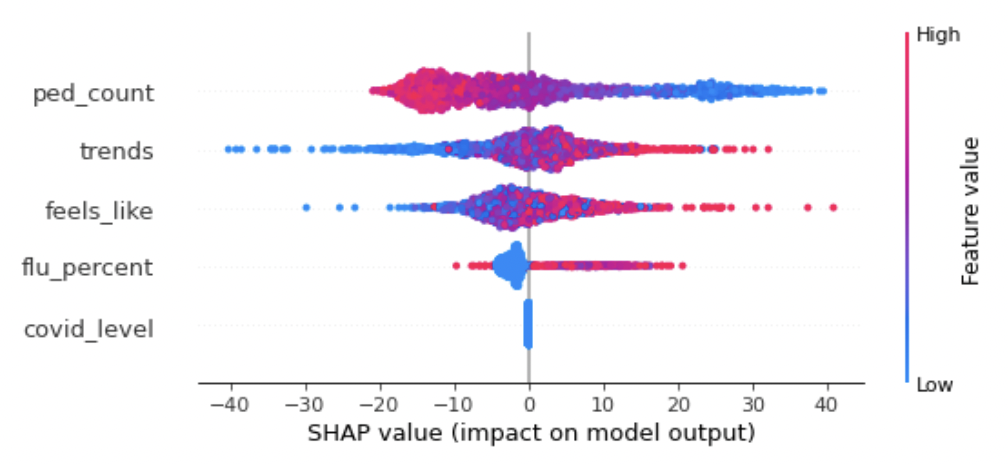}} & 
\subfloat(b){%
\includegraphics[scale=0.3]{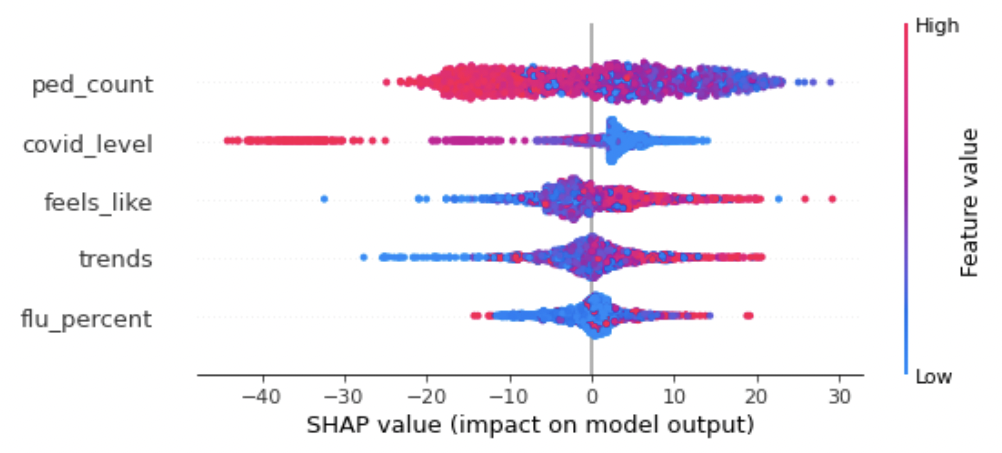}
} \\
\subfloat(c){%
\includegraphics[
scale=0.2]{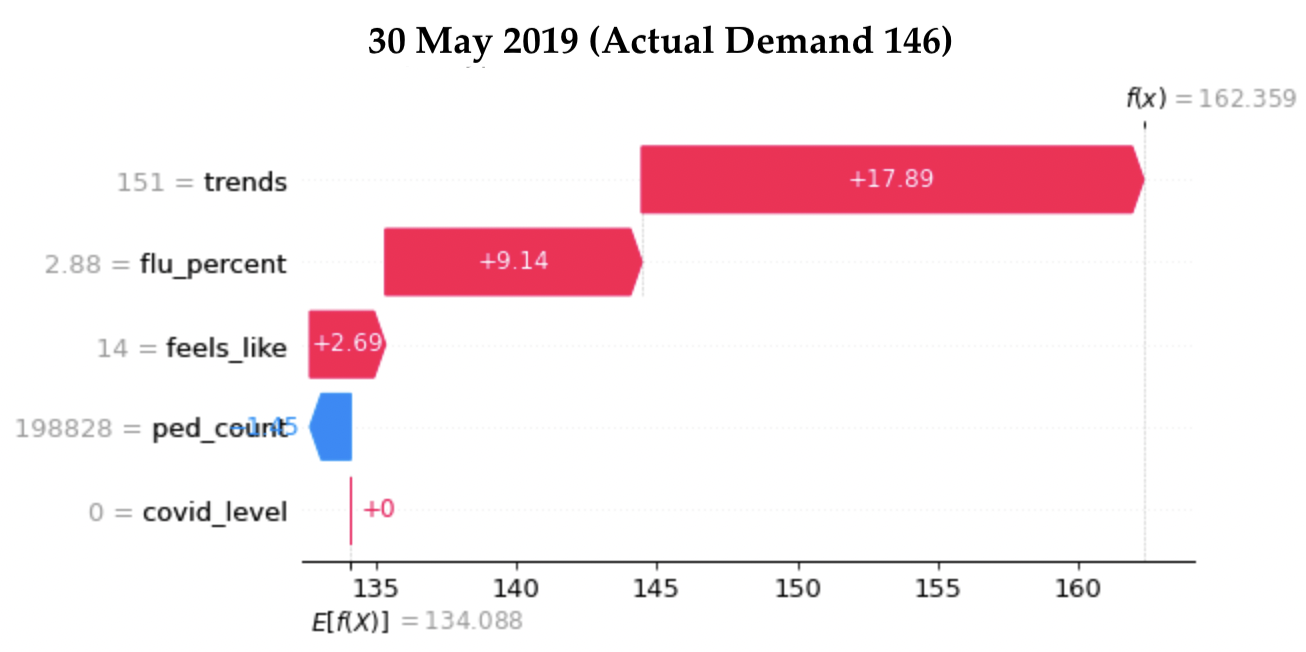}} & 
\subfloat(d){%
\includegraphics[
scale=0.2]{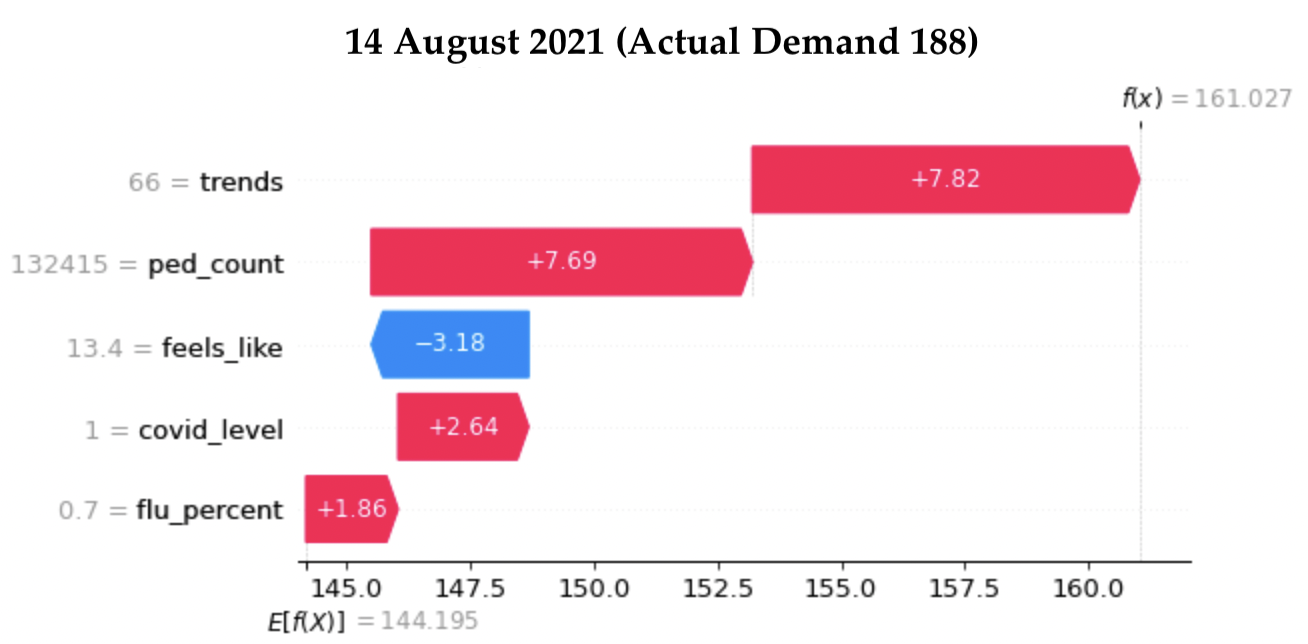}
} \\
\subfloat(e){%
\includegraphics[
scale=0.2]{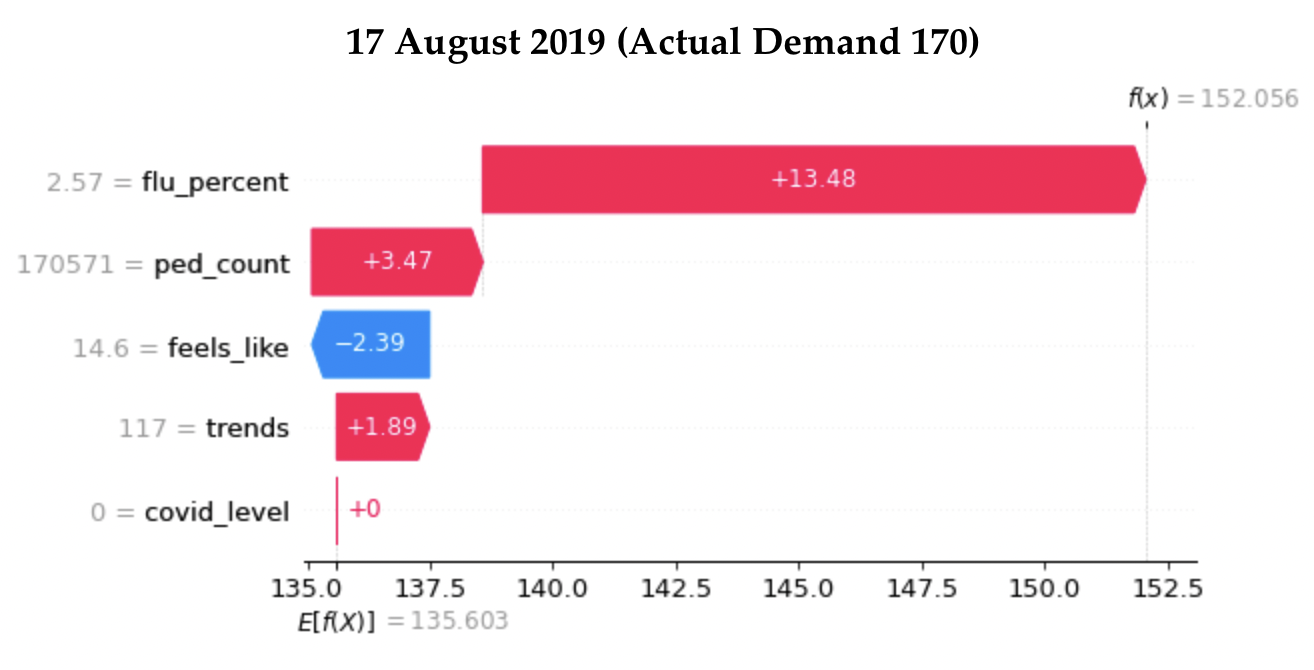}} & 
\subfloat(f){%
\includegraphics[
scale=0.2]{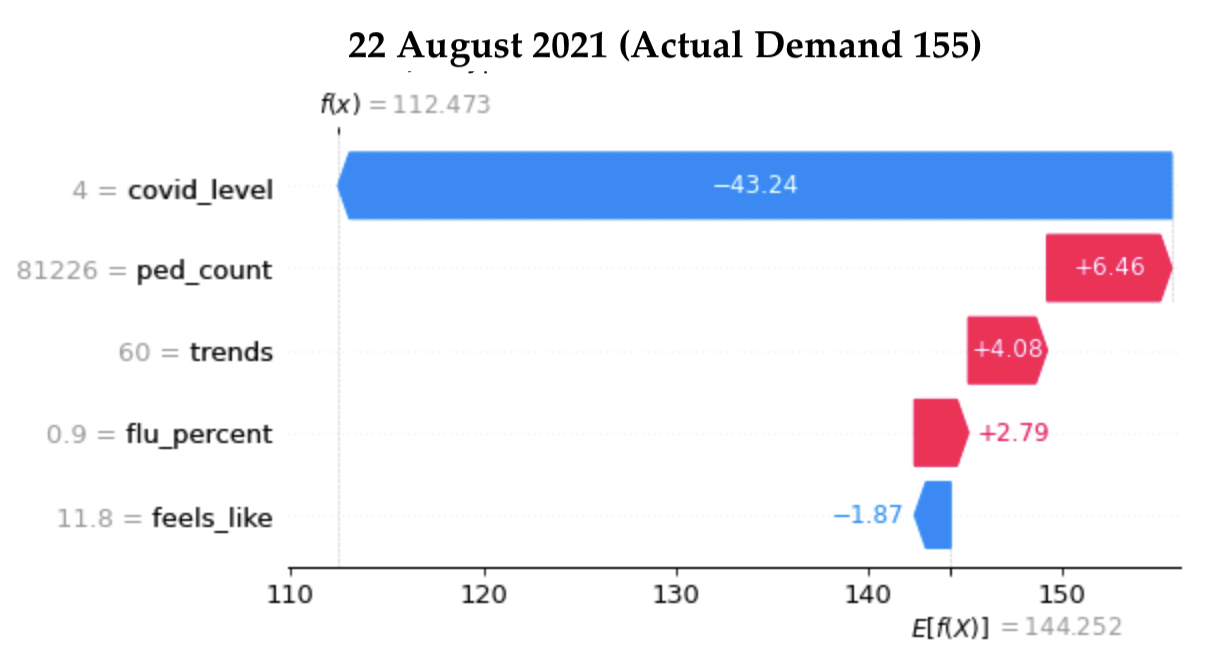}
} \\
\end{tabular}
\caption{SHAP proxy-only model graphs for 2019 (Left) and 2021 (Right). (a) and (b) depict feature importance summary plots. (c) - (f) show explanations of forecasts for specific dates together with the observed values.)}
\label{Fig:Shap_proxy_only}
\end{figure}

Here, we revisit the proxy-only models to study their effects and utility, which we conduct with the removal of the autoregressive features. Figures \ref{Fig:Shap_proxy_only}a-f depict the behaviour of these models. The same analysis approach is followed as previously by using data up to 2019 and 2021. Again, the first two Figures \ref{Fig:Shap_proxy_only}a and \ref{Fig:Shap_proxy_only}b depict the interpretability of the models. In both years, it can be seen that pedestrian foot traffic is a significant driver of the forecasts being negatively correlated with the forecasted patient flows. Google trends and 'feels like' data similarly exert a  positively correlated influence across both years; however, for 2021 the COVID-19 Alert Level becomes the second most important feature with a negatively correlated impact on the patient flows. 

Figures \ref{Fig:Shap_proxy_only}c and \ref{Fig:Shap_proxy_only}e  depict the explainability of the forecasts for two separate dates in 2019. Both explanatory figures demonstrate the strong influence of the prevailing influenza conditions as a strong variable for this period, while Google trends searches are the strongest for the former. For 2021, we observe that Google trends search results, as well as pedestrian foot traffic, are influential on both dates in Figures \ref{Fig:Shap_proxy_only}d and \ref{Fig:Shap_proxy_only}f; however, the COVID-19 Alert Level is an overwhelmingly influential variable for the latter in driving the forecasts down due to the highest prevailing pandemic level of 4 occurring in Figure \ref{Fig:Shap_proxy_only}f. The different magnitudes of effect that the COVID-19 Alert Level is explained by Figure \ref{Fig:Shap_proxy_only}b, where we see that high alert levels exert a considerably larger effect on the forecast than lower alert levels exert on increasing the forecasts.

\begin{figure}[hbt]
\begin{tabular}{cc}
\subfloat(a){%
\includegraphics[scale=0.35]{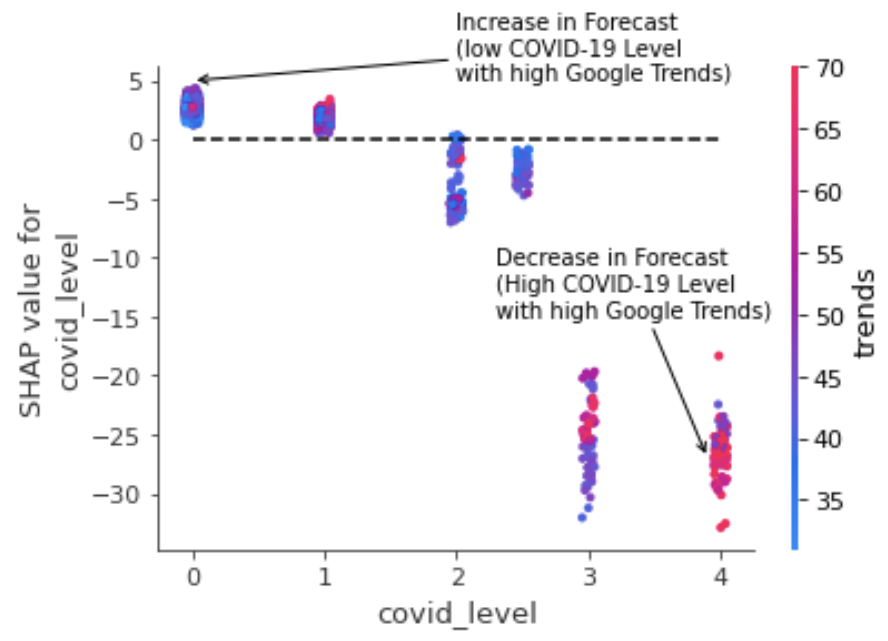} 
} & 
\subfloat(b){%
\includegraphics[scale=0.32]{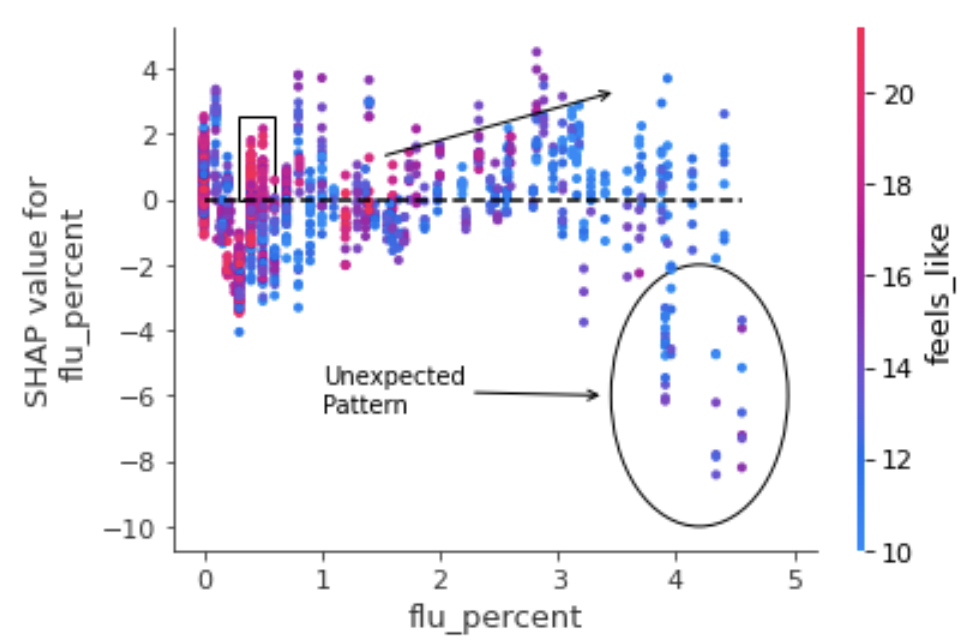}
} \\
\subfloat(c){%
\includegraphics[scale=0.35]{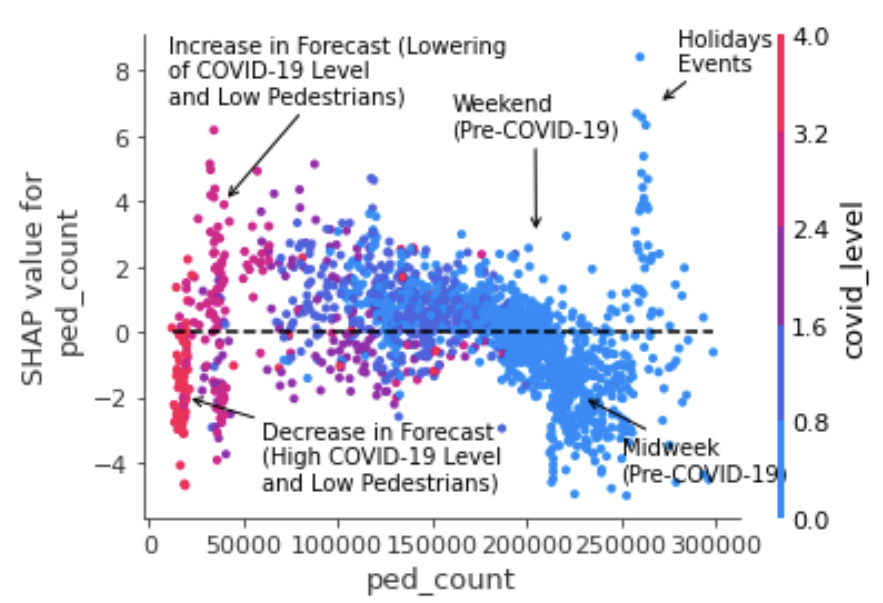} } & 
\subfloat(d){%
\includegraphics[scale=0.35]{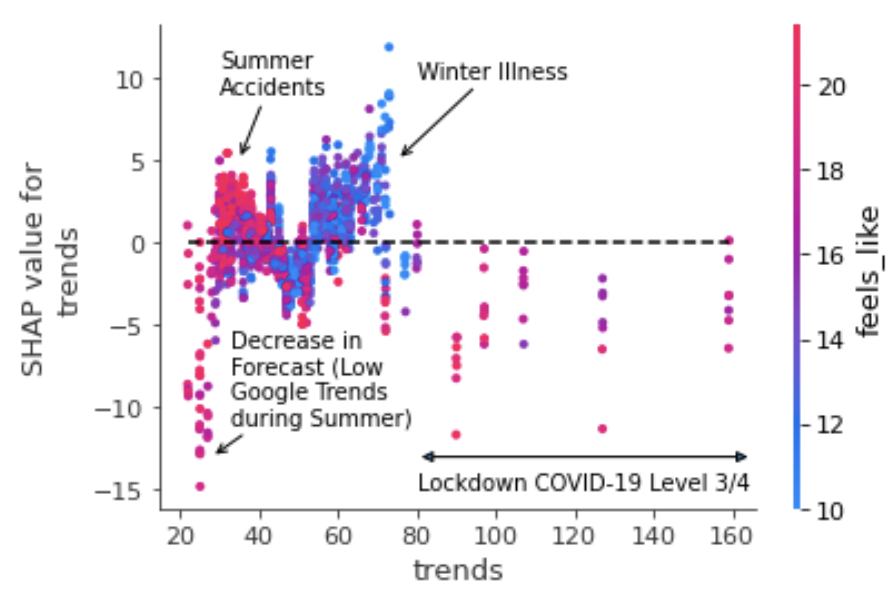}
} \\
\end{tabular}
\caption{SHAP feature dependence plots showing interaction effects between various pairs of proxy variables.}
\label{Fig:Shap_dependence}
\end{figure}

We conclude with a high-level analysis of the model's behaviour to draw out further insights, but this time considering how pairs of features interact in order to influence the final forecasts. We depict a selection of these interactions in Figures \ref{Fig:Shap_dependence}a to \ref{Fig:Shap_dependence}d focusing primarily on proxy features. Each plot contains a horizontal line which indicates the threshold representing a transition from increasing and decreasing effects that pairs of features have on the final forecast.   

Figure \ref{Fig:Shap_dependence}a shows how the effect on the final forecast changes as the COVID-19 Alert Level increases (x-axis) and as the value for Google Trends search terms rises (colour gradient on the right y-axis). The interaction of low COVID-19 Alert Level values and high Google Trends search terms exerts the strongest effect on raising the forecasts. The forecasts are significantly downgraded as the COVID-19 Alert Level values increase and the lockdown measures take effect but at the same time, some of the highest Google Trends search terms regarding respiratory terms are encountered during the highest lockdown conditions\footnote{The greatest values indeed coincided with the first and the highest COVID-19 Alert Level announced in March 2020}. 

Several insights can be extracted from Figure \ref{Fig:Shap_dependence}b depicting the interaction between the reported increase in influenza prevalence and the weather variable. Firstly, it can be observed that influenza levels between 0.5\% and 1\% in conjunction with high weather values tend to correspond with positive effects on the forecasts. This can be explained by summer conditions where there is an uptick in sport-related injuries and general outdoor accident-based cases. However, influenza levels below 0.5\% tend to be inconsistent. As the influenza prevalence increases from 1\% to 3\%, this has a positive effect on forecasts as would be expected during the winter months, and can be confirmed by the fact that most of the weather-perception values are downward trending. An unexpected pattern can be seen as influenza prevalence levels increase beyond 3\% where counter-intuitively a mixed, and more negative effect is exerted on the forecasts which is contrary to the reported experience of the clinic's staff\footnote{The negative effect of high influenza prevalence may be due to uncertainty in the data itself which raises some questions about the reliability of this feature since it is incomplete with the data being available from 2018 onward. It may also be explained by the fact that this feature is a nationwide indicator and influenza incidence is likely to be different in colder regions of the country than in Auckland. The fact that the highest values of flu prevalence coincide with high weather values as negative drivers of forecasts may be suggestive that this period corresponds with the earlier beginning of spring in Auckland than in other regions, which leads to a sharp actual decline in respiratory related illness.}.  

Meanwhile, in Figure \ref{Fig:Shap_dependence}c it is observable that as the pedestrian traffic increases, there is a generally a decreasing effect on the forecasts\footnote{This is explainable by the fact that the pedestrian traffic peaks midweek when the patient arrivals are at their lowest. Meanwhile, the pedestrian traffic is lowest in the CBD during the weekend when patient demand is typically at its highest.} until the highest pedestrian traffic values are reached, at which point, the behaviour of the models becomes more erratic, possibly due to public holidays and large events. The highest  COVID-19 Alert Level values together with the ensuing lowest pedestrian traffic values, aggressively decrease the forecasts as expected. The highest positive effects on the forecasts are seen in data points where the pedestrian traffic is low but the COVID-19 Alert Level is less than 4 which corresponds with the most stringent lockdown mandates. This also corresponds with the loosening of the lockdown restrictions which in turn trigger increased clinic visits due to the pent-up demand during the highest lockdown conditions. 

In the last illustration, Figure \ref{Fig:Shap_dependence}d depicts the effects of increasing Google Trends search terms with respect to changes to weather conditions. We see that some of the highest  Google Trends search values (from 80 onward) correspond with the strongest negative forecasting effects. We can correlate this result with Figure \ref{Fig:Shap_dependence} which shows that some of the highest Google Trends values occurred during the highest lockdown measures thus explaining why these values are exerting a strong negative effect on the forecasts. Therefore, it can be surmised that unusually high Google Trends search values from 80 onward are exceptional as they are attributable to severe lockdown conditions. Meanwhile, Google Trends search values ranging from 50 to 80 can be interpreted as belonging to the winter months and represent typical patterns that are confirmed by the low weather temperatures for this range. The effect of the summer months and the lowest Google Trends search values can also be observed where the strongest negative effects on the forecast values are exerted. 

\section{Discussion}\label{sec12}

A comprehensive empirical investigation was conducted using a broad set of models and features which achieved considerably improved accuracies over the benchmark approaches as well as competitive accuracies with respect to prior studies. In addressing the study's first research question (RQ1), we find that the ensemble-based Voting method consistently outperformed all other candidate algorithms for this setting. The approach successfully combined a mixture of both machine learning and statistical techniques in order to leverage the advantages of both. 

The abrupt onset of the concept drift caused by the COVID-19 pandemic conditions degraded the forecasting accuracies initially. Nonetheless, the experiments indicated that the suite of novel real-time proxy variables were effective at adjusting the models to the new and evolving drivers of patient flows. Not only were the proposed features useful at triggering adaptations to the disruptions of historic patterns, but they were also effective at improving the model forecasts in general (RQ2). Indeed, the experiments showed that reasonable models could be generated using only proxy variables without reliance on autoregressive feature values. The study confirmed that the COVID-19 Alert Level feature was particularly effective at forecasting patient volumes during disruptive periods, while determining the effectiveness of the Google search terms, pedestrian foot traffic as well as weather data. While the reported prevalence of influenza prevalence was a useful feature and contributed to some degree in improving forecasts, it was less reliable than the other proxy features likely due to its incompleteness. The implications of this finding is that accurate forecasting of patient flows should remain possible using these features in the event of future pandemic outbreaks.

A particularly novel contribution of this study was the extensive use of XAI tools in order to expose the internals of the forecasting models. The use of the SHAP technique to achieve global interpretability of the models was highlighted and proved effective in depicting the various effects that features and their values bear on the final forecasts (RQ3). Additionally, at the local level, the technique demonstrated its ability to clearly explain the drivers of the forecasted values for individual days. Meanwhile, the dependence plots yielded extra insights into the interactions of various features and how they collectively influence the final forecasts. This also enabled the validation of the models against domain knowledge to take place.

\section{Conclusion}\label{sec13}

Patient flows in Urgent Care Clinics (UCCs) and Emergency Departments (EDs) have been experiencing increasing pressure as the volumes have grown and have become less predictable using \textit{ad hoc} approaches. The ability to accurately forecast patient arrivals in these contexts, and to understand better what some of the drivers of demand are, is important for the efficient and effective functioning of these healthcare providers by enabling them to respond faster and achieve more optimal human resourcing.

There are many factors which affect patient arrivals and randomness accounts for some of this. Autoregressive variables as well as calendar and meteorological indicators have already been explored in prior research and have been found useful for estimating patient flows. This study goes further by also considering additional quasi-real-time variables like Google search terms, pedestrian traffic and prevailing incidence levels of influenza, alongside COVID-19 Alert Level indicators to improve the forecast accuracy. 

This study makes a unique contribution in several respects. Not only does this study integrate a wide variety of new features for forecasting patient flows, but it also considers both the effects that the concept drift triggered by the recent pandemic conditions had on the forecast accuracies, and the means by which the models can rapidly adapt to the changing context in order to learn and recover. This research also ventured beyond the standard studies using machine learning methods in this domain, by utilising tools from the eXplainable AI field in order to expose the internals of the models, thus achieving both the high-level interpretability of the models and the explainability of their individual forecasts.

Our research determined that the Voting ensemble-based method performed most reliably in this setting and the final accuracies were competitive with those from prior studies. While the autoregressive and calendar features were important, the experiments indicated that the prevailing COVID-19 Alert Level feature together Google search terms and pedestrian traffic were effective at generating accurate forecasts.

\section*{Acknowledgments}
The authors would like to thank ShoreCare and their staff for both their ongoing assistance in working with the data, and provisioning the datasets and for ultimately enabling this research to take place.

\section*{Competing interests}
The authors declare that they have no competing interests.

\section*{Availability of data and materials}
The data for this study cannot be made available due to its sensitivity. 

\section*{Funding}
Not applicable.

\section*{Author's contributions}
The authors contributed equally to this research. PM performed data acquisition, cleaning and the writing of all software for conducting the experiments and the implementation of the models. TS conceptualised the aims and the direction of the study and wrote the initial draft. Both authors analysed the results as well as contributed to the final manuscript, and approve its submission for publication.


\urlstyle{same}

\appendix

\section{Detailed Model Accuracies and Statistical Test Results}\label{appendix:improvements}

\subsection*{Quantifying improvements of proposed methods versus the benchmark.}
Tables \ref{table:Perc_Imp_S} and \ref{table:Perc_Imp_N} show the improvements achieved by the top seven proposed methods using all features over the benchmark as a percentage of MAPE. The results are broken down by year.

\begin{table}[hbt!]
\parbox{.45\linewidth}{
\tiny
\caption{Percentage improvement of proposed models over the benchmark for Clinic 1 across all years.} 
\tiny
\centering
\setlength\tabcolsep{1.5pt}
\begin{tabular}{lccccccc}
\hline
 & 2017  & 2018 & 2019 & 2020 
& 2021  & 2022 & Overall  \\
\hline
Prophet & 26.8 & 29.8 & 25.5 & 27.7 & 26.1 & 3.3 & 25.0 \\   Random Forest & 17.1 & 26.2 & 27.8 & 38.5 & 29.7 & 22.2 & 27.3 \\   CatBoost & 13.6 & 27.6 & 26.5 & 43.5 & 30.5 & 25.3 & 28.1 \\   Gradient Boosting & 18.4 & 26.9 & 28.5 & 43.1 & 31.3 & 24.6 & 29.2 \\   Stacking & 25.7 & 30.6 & 30.1 & 40.5 & 31.0 & 19.1 & 30.5 \\   Averaging & 24.6 & 31.3 & 32.1 & 42.0 & 34.5 & 25.4 & 32.2 \\   Voting & 25.5 & 31.0 & 32.3 & 41.2 & 36.1 & 27.2 & 32.7
\\
\hline
\end{tabular}\label{table:Perc_Imp_S}
}
\hfill
\parbox{.45\linewidth}{
\tiny
\caption{Percentage improvement of proposed models over the benchmark for Clinic 2 across all years.} 
\tiny
\centering
\setlength\tabcolsep{1.5pt}
\begin{tabular}{lccccccc}
\hline
 & 2017  & 2018 & 2019 & 2020 
& 2021  & 2022 & Overall  \\
\hline
Prophet & 25.8 & 12.0 & 26.2 & -25.0 & 29.6 & -15.1 & 11.2 \\   Gradient Boosting & 21.6 & 12.2 & 28.7 & -0.7 & 38.0 & 17.2 & 19.7 \\   Random Forest & 23.8 & 13.0 & 27.3 & 2.0 & 36.1 & 19.1 & 20.3 \\   CatBoost & 23.0 & 12.0 & 26.9 & 5.1 & 38.1 & 14.0 & 20.4 \\   Stacking & 27.5 & 17.0 & 33.7 & -2.6 & 37.9 & 13.6 & 21.9 \\   Averaging & 29.0 & 16.8 & 34.0 & 1.1 & 39.4 & 16.1 & 23.4 \\   Voting & 29.8 & 16.6 & 34.6 & -1.2 & 40.3 & 17.2 & 23.4

\\
\hline
\end{tabular}\label{table:Perc_Imp_N}
}
\end{table}

\subsection*{TheilU statistical test on the significance of model improvements over the benchmark.}

Tables \ref{table:TheilU_S} and \ref{table:TheilU_N} display the results of the TheilU statistical test showing the detailed improvements attained by the top seven proposed methods using all features over the benchmark. The test is performed across all years as a summary as well as broken down by year for each clinic. The tables show that all the values for the Voting model are below 1, indicating an improvement over the in-house method, apart from one instance in 2020 for Clinic 2, where the performance was comparable.

\begin{table}[hbt!]
\parbox{.45\linewidth}{
\caption{TheilU Statistic for Clinic 1} 
\tiny
\centering
\setlength\tabcolsep{1.5pt}
\begin{tabular}{lccccccc}
\hline
 & 2017  & 2018 & 2019 & 2020 
& 2021  & 2022 & Overall  \\
\hline
Random Forest & 0.576 & 0.623 & 0.635 & 0.99 & 0.894 & 0.989 & 0.762 \\   Prophet & 0.507 & 0.584 & 0.612 & 0.927 & 0.873 & 1.417 & 0.756 \\   Gradient Boosting & 0.566 & 0.622 & 0.63 & 0.863 & 0.872 & 0.965 & 0.721 \\   CatBoost & 0.598 & 0.599 & 0.629 & 0.85 & 0.866 & 0.959 & 0.719 \\   Stacking & 0.516 & 0.573 & 0.596 & 0.845 & 0.865 & 1.127 & 0.706 \\   Averaging & 0.525 & 0.565 & 0.588 & 0.833 & 0.816 & 1.000 & 0.685 \\   Voting & 0.516 & 0.559 & 0.581 & 0.821 & 0.795 & 0.956 & 0.671

\\
\hline
\end{tabular}\label{table:TheilU_S}
}
\hfill
\parbox{.45\linewidth}{
\caption{TheilU Statistic for Clinic 2} 
\tiny
\centering
\setlength\tabcolsep{1.5pt}
\begin{tabular}{lccccccc}
\hline
 & 2017  & 2018 & 2019 & 2020 
& 2021  & 2022 & Overall  \\
\hline
Prophet & 0.774 & 0.76 & 0.591 & 1.584 & 1.064 & 1.133 & 1.045 \\   Stacking & 0.725 & 0.693 & 0.56 & 1.167 & 0.888 & 0.982 & 0.855 \\   Random Forest & 0.743 & 0.73 & 0.631 & 1.026 & 0.962 & 0.969 & 0.845 \\   Averaging & 0.703 & 0.689 & 0.566 & 1.052 & 0.871 & 0.967 & 0.815 \\   Gradient Boosting & 0.736 & 0.732 & 0.626 & 0.894 & 0.916 & 1.043 & 0.810 \\   Voting & 0.692 & 0.684 & 0.558 & 1.009 & 0.886 & 0.947 & 0.800 \\   CatBoost & 0.754 & 0.737 & 0.63 & 0.836 & 0.872 & 1.069 & 0.796
\\
\hline
\end{tabular}\label{table:TheilU_N}
}
\end{table}

\subsection*{Accuracies of the proxy-only model}

Table \ref{table:Proxy_Only_Metrics_S} shows the detailed MAPE accuracies of the proxy-only models for Clinic 1 contrasted with univariate ARIMA and Prophet models. The univariate models outperform the proxy-only models in the period leading up to the pandemic, but subsequently the Voting proxy-only models shows general improvements over the univariate methods as the concept drift takes effect from 2020 onward.

\begin{table*}[hbt!]
\caption{MAPE accuracies for the proxy-only models for Clinic 1} 
\tiny
\centering
\setlength\tabcolsep{1.5pt}
\begin{tabular}{lcccccccccccccc}
\hline
 & \multicolumn{2}{c}{2017} 
 & \multicolumn{2}{c}{2018} 
 & \multicolumn{2}{c}{2019} 
 & \multicolumn{2}{c}{2020} 
 & \multicolumn{2}{c}{2021} 
 & \multicolumn{2}{c}{2022} 
 & \multicolumn{2}{c}{Overall Mean} 
\\
& MAPE & R & MAPE & R & MAPE & R & MAPE & R 
& MAPE & R & MAPE & R & MAPE & R
\\
\hline
CatBoost & 16.9 & 4.5 & 14.0 & 4.1 & 14.0 & 4.1 & 16.5 & 3.7 & 12.8 & 3.6 & 14.0 & 3.8 & 14.8  & 4.0 \\   Random Forest & 16.3 & 4.2 & 14.0 & 4.2 & 14.3 & 4.3 & 16.3 & 3.7 & 13.0 & 3.8 & 13.3 & 3.3  & 14.6 & 4.0 \\   ARIMA & 9.7 & 2.0 & 9.7 & 2.5 & 9.0 & 2.3 & 13.1 & 2.6 & 10.8 & 2.7 & 10.5 & 2.4 & 10.5 & 2.4 \\   Prophet & 8.5 & 1.7 & 8.4 & 1.9 & 8.4 & 2.1 & 13.3 & 2.8 & 10.2 & 2.6 & 14.2 & 3.5 & 10.2 & 2.3 \\   Voting & 11.0 & 2.6 & 9.7 & 2.3 & 9.7 & 2.3 & 12.4 & 2.1 & 10.0 & 2.2 & 11.3 & 2.0 & 10.6 & 2.3

\\
\hline
\end{tabular}

\label{table:Proxy_Only_Metrics_S}
\end{table*}

\section{Proxy feature values versus patient flows}\label{appendix:proxy_vars}
\begin{figure}[hbt!]
\centering
\includegraphics[scale=0.3]{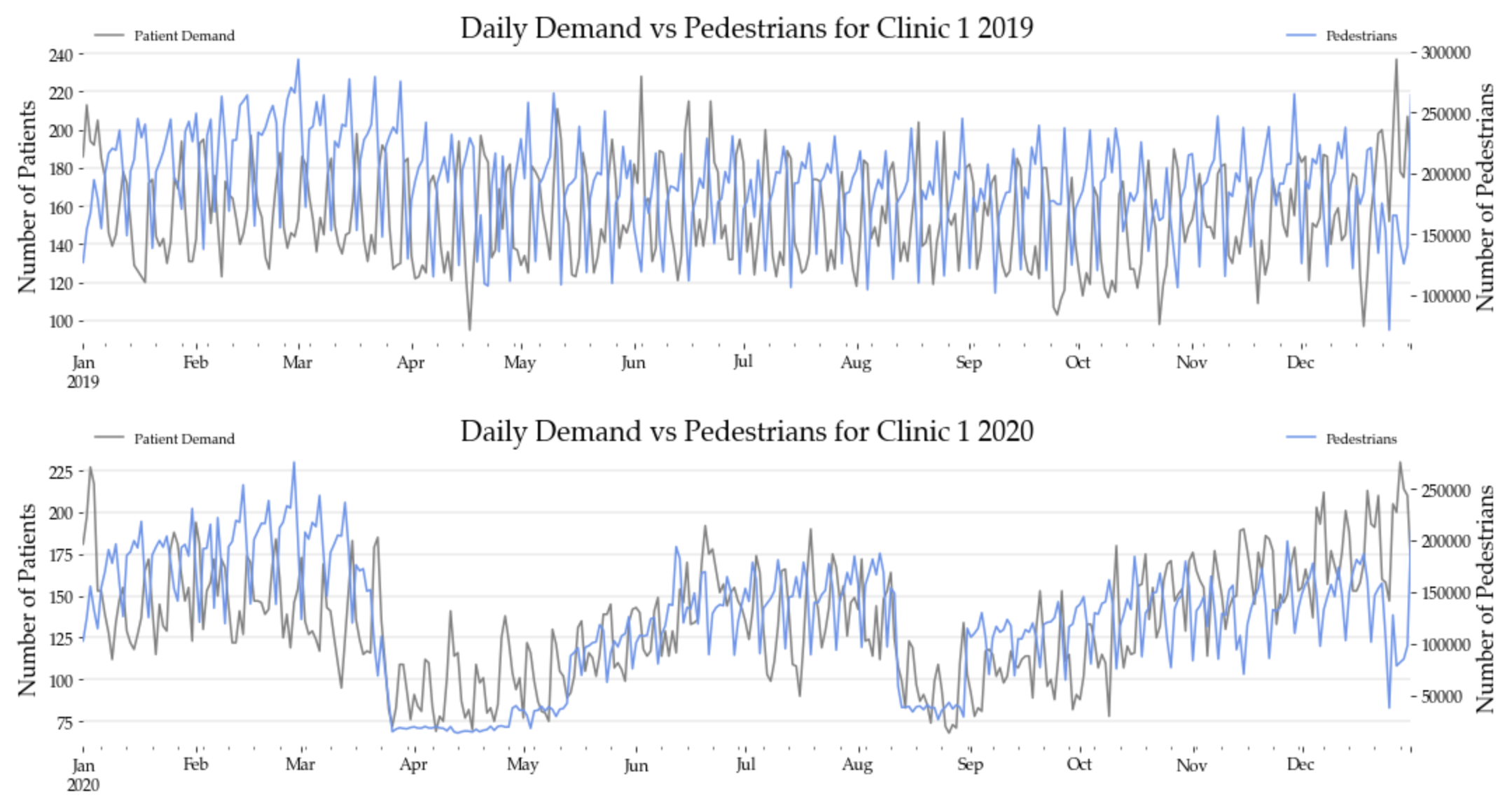}
\caption{Pedestrian Traffic Proxy Feature for 2019 and 2020.}
\label{figure:proxy_ped}
\end{figure}

\begin{figure}[hbt]
\centering
\includegraphics[scale=0.3]{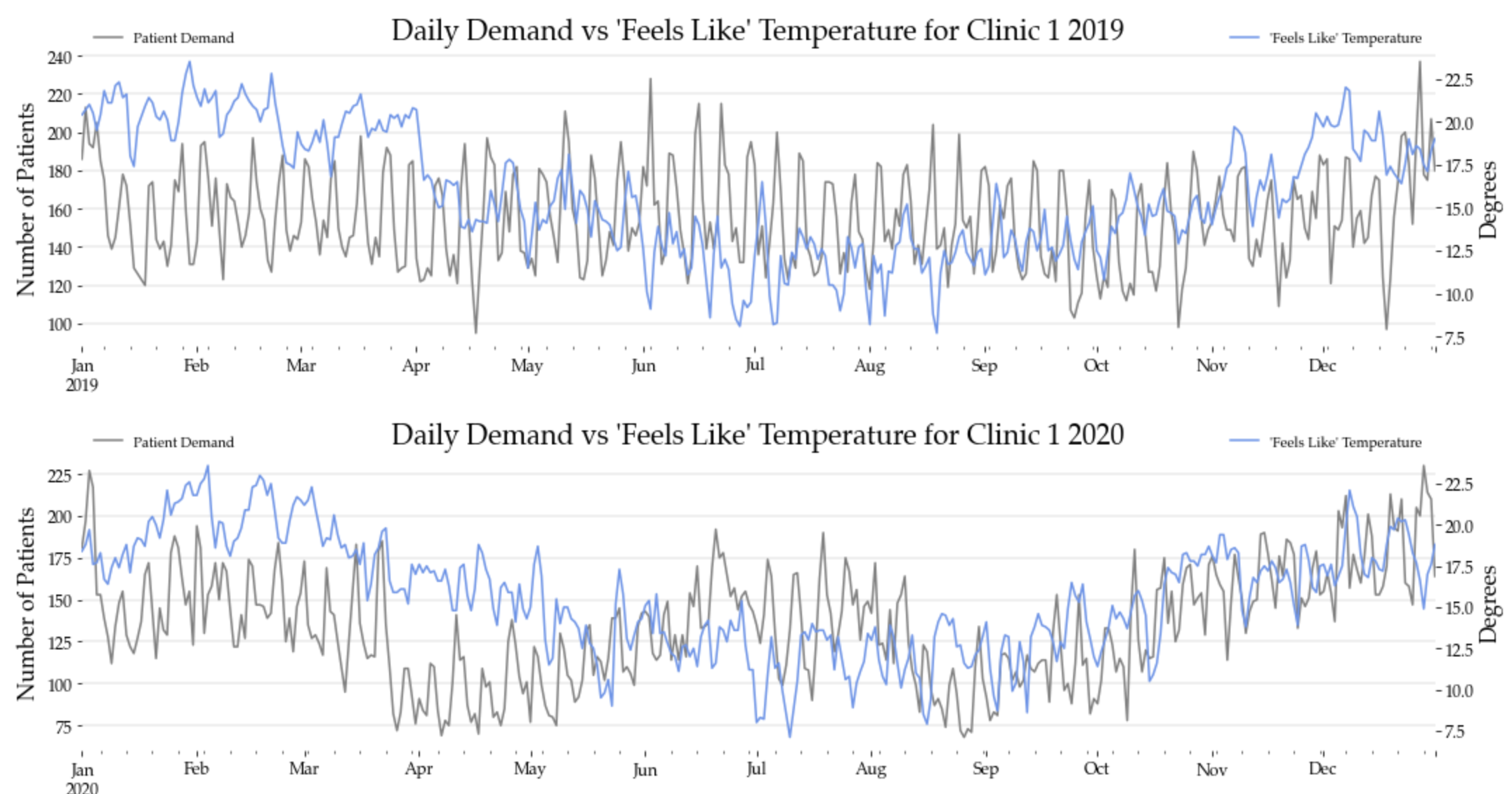}
\caption{'Feels Like' Temperature Proxy Feature for 2019 and 2020.}
\label{figure:proxy_temp}
\end{figure}

\begin{figure}[hbt]
\centering
\includegraphics[scale=0.3]{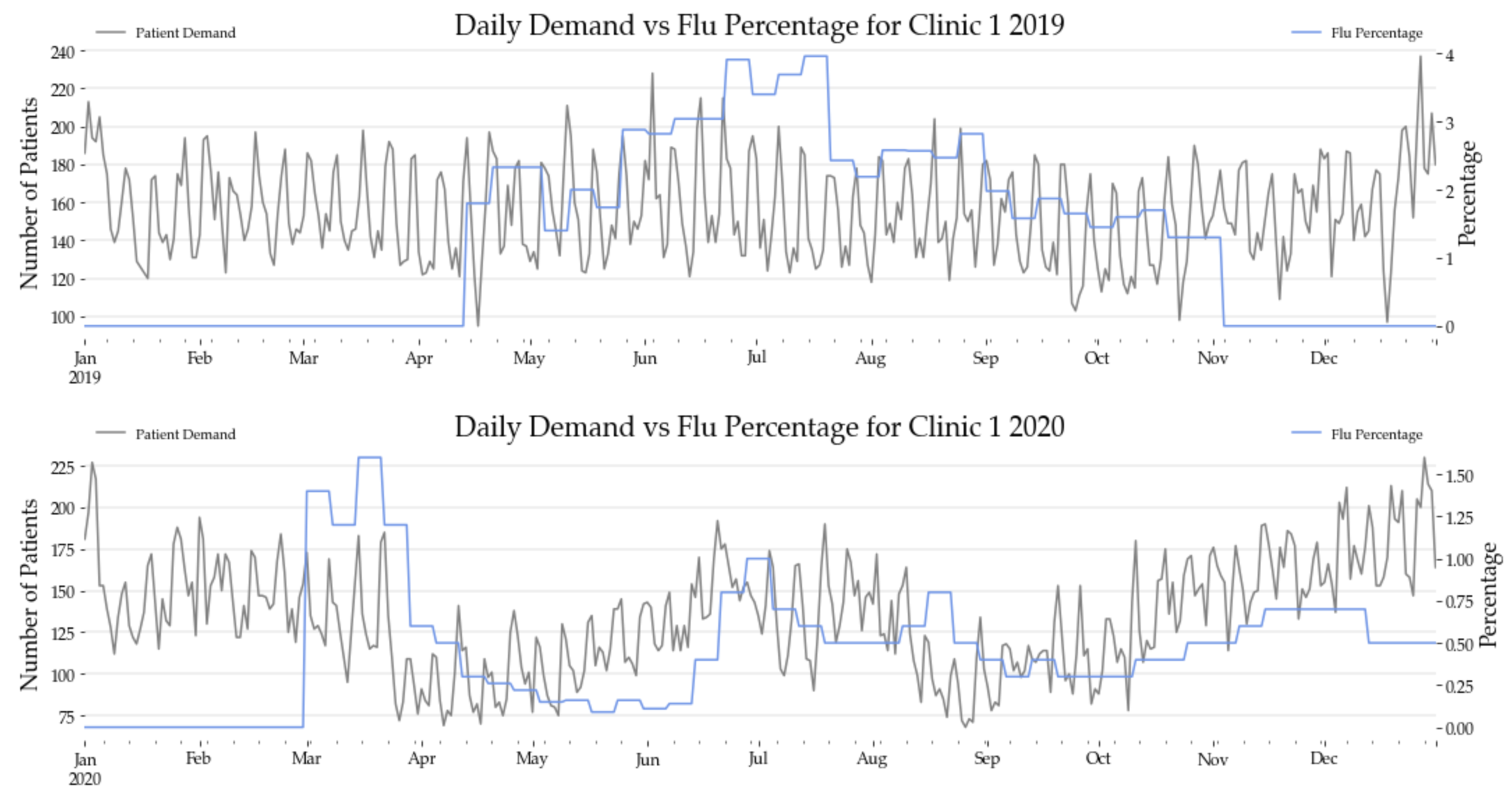}
\caption{Flu Tracker Proxy Feature for 2019 and 2020.}
\label{figure:proxy_flu}
\end{figure}

\begin{figure}[hbt]
\centering
\includegraphics[scale=0.42]{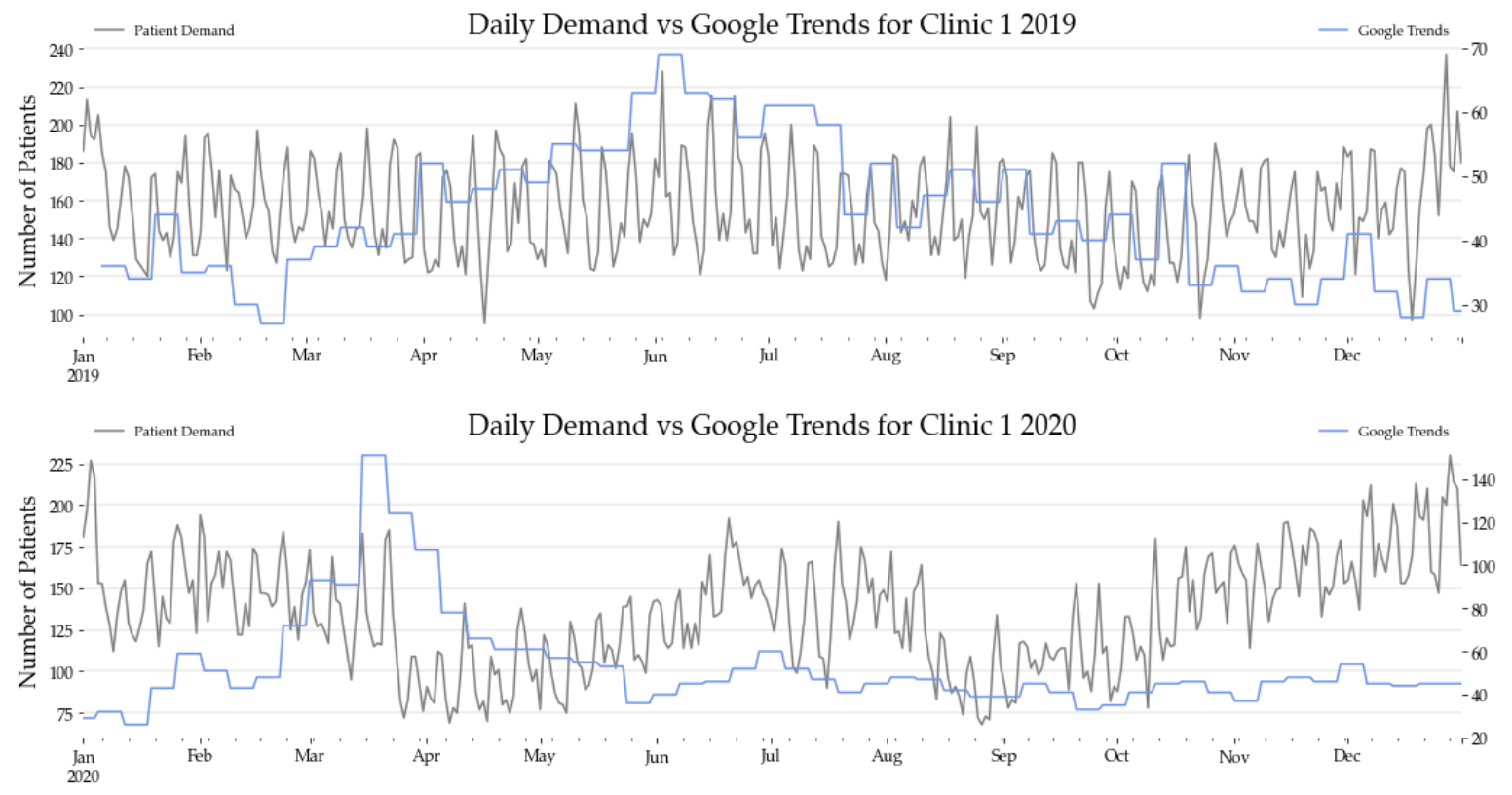}
\caption{Google Trends Proxy Feature for 2019 and 2020.}
\label{figure:proxy_trends}
\end{figure}

Figures \ref{figure:proxy_ped} to \ref{figure:proxy_trends} highlight the correlation between the patient arrivals per day for Clinic 1 against the proxy variable values for 2019 and 2020, contrasting trends over stable and concept drift periods.   

\end{document}